\documentclass[10pt,twocolumn,letterpaper]{article}

\usepackage{iccv}
\usepackage{times}
\usepackage{epsfig}
\usepackage{graphicx}
\usepackage{amsmath}
\usepackage{amssymb}
\usepackage{multirow}
\usepackage{verbatim}
\usepackage{caption}
\usepackage{float} 
\usepackage{subcaption}
\usepackage{arydshln}
\usepackage{bm}
\usepackage{color}

\usepackage[pagebackref=true,breaklinks=true,letterpaper=true,colorlinks,bookmarks=false]{hyperref}

\iccvfinalcopy 

\ificcvfinal\fi

\begin{document}

\title{Hierarchical Prior Mining for Non-local Multi-View Stereo}

\author{Chunlin Ren$^1$\;
Qingshan Xu$^2$\;   
Shikun Zhang$^1$\;
Jiaqi Yang$^1$\thanks{Corresponding author}\\
$^1$ Northwestern Polytechnical University\;                                                
$^2$ Nanyang Technological University\;
}

\maketitle
\ificcvfinal\thispagestyle{empty}\fi

\begin{abstract}
As a fundamental problem in computer vision, multi-view stereo (MVS) aims at recovering the 3D geometry of a target from a set of 2D images. Recent advances in MVS have shown that it is important to perceive non-local structured information for recovering geometry in low-textured areas. In this work, we propose a Hierarchical Prior Mining for Non-local Multi-View Stereo (HPM-MVS). The key characteristics are the following techniques that exploit non-local information to assist MVS: 1) A Non-local Extensible Sampling Pattern (NESP), which is able to adaptively change the size of sampled areas without becoming snared in locally optimal solutions. 2) A new approach to leverage non-local reliable points and construct a planar prior model based on K-Nearest Neighbor (KNN), to obtain potential hypotheses for the regions where prior construction is challenging. 3) A Hierarchical Prior Mining (HPM) framework, which is used to mine extensive non-local prior information at different scales to assist 3D model recovery, this strategy can achieve a considerable balance between the reconstruction of details and low-textured areas. Experimental results on the ETH3D and Tanks \& Temples have verified the superior performance and strong generalization capability of our method. Our code will be released.
\end{abstract}

\section{Introduction}

Multi-View Stereo (MVS) is one of the key problems in the field of 3D computer vision, which contributes greatly to virtual reality~\cite{bruno20103d}, 3D object recognition~\cite{hejrati2014analysis}, and autonomous driving~\cite{ma2019accurate}. 
The objective of MVS method is to reconstruct the 3D geometry of a target from a series of images along with camera parameters. Over the past few years, the emergence of numerous datasets~\cite{jensen2014large, schops2017multi, yao2020blendedmvs} has tremendously contributed to the continuous advancement of MVS methods~\cite{furukawa2009accurate, shan2013visual, galliani2015massively}. However, influenced by low-textured areas, illumination variation and other factors~\cite{aanaes2016large}, MVS is still a challenging problem.

\begin{figure}[htbp]
    \centering
    \begin{subfigure}{0.49\linewidth}
		\centering
		\includegraphics[width=1\linewidth]{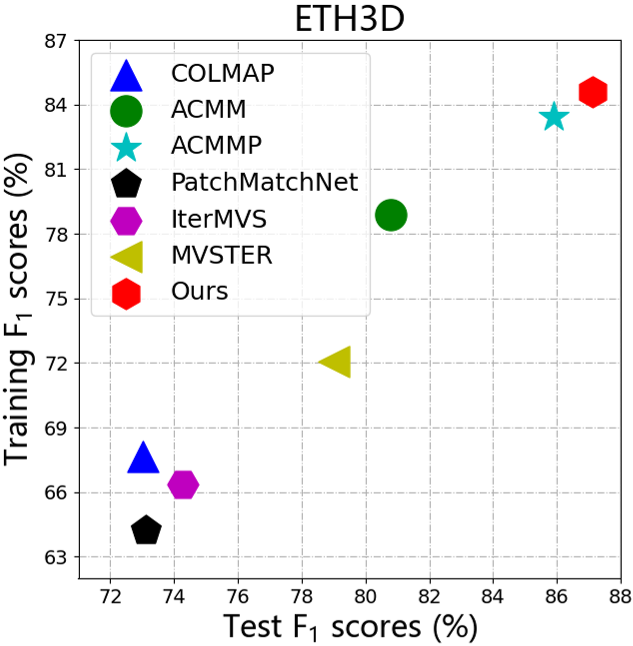}
	\end{subfigure}
        \centering
	\begin{subfigure}{0.49\linewidth}
		\centering
		\includegraphics[width=1\linewidth]{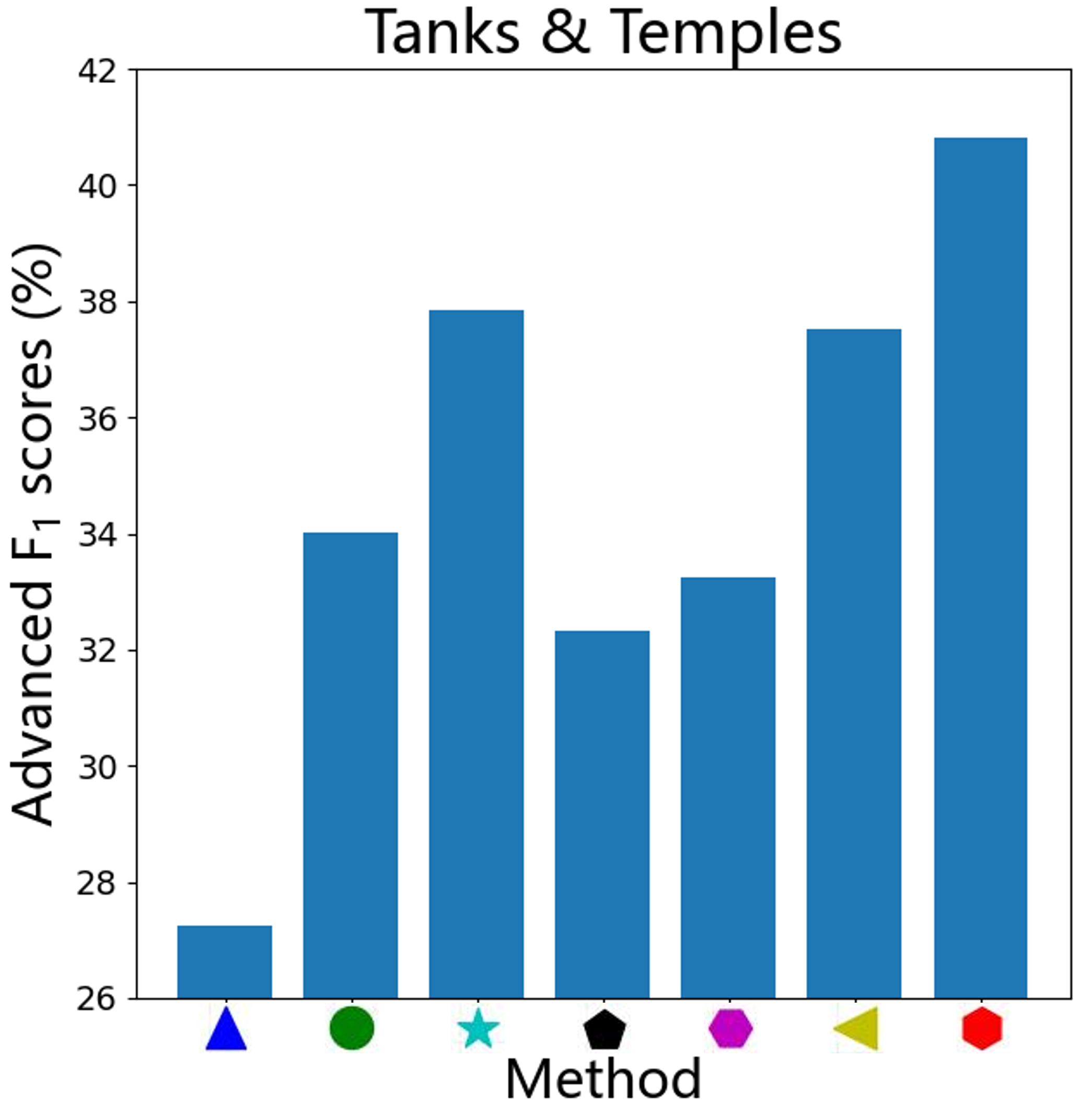}
	\end{subfigure}
    \caption{$F_1$ score ($\uparrow$) comparisons with SOTA traditional~\cite{schonberger2016pixelwise, xu2019multi, xu2022multi} and learning-based~\cite{wang2021patchmatchnet, wang2022itermvs, wang2022mvster} MVS methods on ETH3D~\cite{schops2017multi} and Tanks \& Temples~\cite{knapitsch2017tanks}.}
    \label{fig: teaser}
\end{figure}

To perform an accurate and robust 3D reconstruction, two different types of methods have been investigated, including learning-based MVS~\cite{yao2018mvsnet, yao2019recurrent, xu2022learning} and traditional MVS~\cite{galliani2015massively, xu2019multi, xu2022multi}. Learning-based MVS implements deep networks to extract high-level features and make predictions. However, it needs a large amount of training and its generalization ability still needs further improvement~\cite{keskar2016large}. Traditional MVS can also be broadly classified into two categories: plane-sweeping-based MVS~\cite{collins1996space, baillard2000plane, gallup2007real} and PatchMatch MVS~\cite{galliani2015massively, schonberger2016pixelwise, xu2019multi}. Although plane-sweeping-based MVS produces good results for sufficiently textured and unoccluded surfaces, it performs poorly for scenes composed of large planar surfaces, while PatchMatch MVS has successfully overcome this restriction. 

PatchMatch MVS usually consists of four steps, including random initialization, hypothesis propagation, multi-view matching cost evaluation and refinement~\cite{xu2022multi}. In essence, it is a process of sampling and verification. The hypothesis propagation samples appropriate hypotheses from neighboring pixels to construct solution space while the multi-view matching cost evaluation defines a criterion to verify the reliability of the sampled hypotheses. Therefore, these two steps play an important role in PatchMatch MVS. Following this popular four-step pipeline, PatchMatch MVS produces remarkable outcomes.

Even so, geometry recovery still suffers a lot from low-textured areas because the previous PatchMatch MVS methods place too much emphasis on local information. Therefore, we exploit non-local information to achieve high-quality reconstruction. We propose a Hierarchical Prior Mining for Non-local Multi-View Stereo (HPM-MVS) method, which defines a novel hypothesis propagation pattern and a novel way for constructing planar prior models to assist multi-view matching cost evaluation.

In terms of hypothesis propagation, sequential propagation~\cite{bailer2012scale, zheng2014patchmatch, schonberger2016pixelwise} and diffusion-like propagation~\cite{galliani2015massively, xu2019multi} are all popular strategies. The previous one only considers the nearest pixels for updating. In contrast, the diffusion-like propagation can update simultaneously based on the neighbor hypotheses around the central point. To further leverage structured region information, ACMH~\cite{xu2019multi} proposes an adaptive checkerboard sampling scheme. Although these previous approaches have greatly improved the efficiency and performance of hypothesis propagation, they both process in the local neighborhood; thus misestimates can only be updated when several better hypotheses in the local area are sampled. Based on the above observations, we first propose the basic method with a Non-local Extensible Sampling Pattern (NESP). Intuitively, a non-local operation removes sampling points around the center point. It allows distant correct pixels to have a greater possibility to contribute to the update of the current pixel. Moreover, the extensible architecture maintains the variable sampling size and can efficiently select suitable candidate hypotheses.

In terms of the multi-view matching cost evaluation, since the photometric consistency usually causes ambiguities during the depth optimization of low-textured areas, many previous methods~\cite{woodford2009global, gallup2010piecewise, xu2020planar} introduce planar prior models to help the evaluation. 
They assume low-textured areas are usually on smooth homogeneous surfaces, and construct a planar prior model at the current scale to assist depth estimation. 
To establish a more robust and more comprehensive planar prior, we employ K-Nearest Neighbor (KNN) to search non-local reliable points and obtain potential hypotheses for the marginal regions where prior construction is difficult. Inspired by the coarse-to-fine strategy, we further design an HPM framework. In a coarse-to-fine manner, the prior knowledge at the early stages is built upon non-local credible sparse correspondences at low-resolution scales, which expands the receptive field and leads to an effective depth estimate for smooth homogeneous surfaces. Subsequently, the prior knowledge at the later stages is constructed at the higher-resolution scales by using previously rectified hypotheses, which restores the depth information of details in the image. In this way, the depth information can be successfully recovered by our HPM architecture. 

Extensive experiments on different competitive datasets verify the effectiveness of our method (Fig.~\ref{fig: teaser}). \textit{The results demonstrate that HPM-MVS has excellent performance, and also holds strong generalization capability to more complex scenes.} In a nutshell, our contributions are three-fold as follows:
\begin{itemize}

    \item We present an NESP module to adaptively determine the number of potential sampling points while avoiding becoming snared in locally optimal solutions.
    \item To construct a more robust and more comprehensive planar prior model, we propose an approach based on KNN to search non-local neighbor reliable hypotheses and construct a planar prior model for marginal areas where prior construction is challenging.
    \item  We design an HPM framework to explore prior knowledge at multiple scales, which efficiently balances out the reconstruction of details and low-textured areas. 
    \end{itemize}

\section{Related Work}
\label{sec:formatting}
\subsection{Traditional Multi-View Stereo}

{\bf \noindent PatchMatch Multi-View Stereo.}  The PatchMatch method was initially proposed by Barnes \etal~\cite{barnes2009patchmatch}, and its primary goal is to quickly identify the approximate nearest neighbor among patches of two images. The extension of PatchMatch idea to MVS was proposed by Shen~\cite{shen2013accurate}. In order to enhance efficiency, several methods~\cite{zheng2014patchmatch, bailer2012scale, wei2014multi} employ the sequential propagation scheme. However, this propagation scheme is still time-consuming. To this end, Galliani \etal~\cite{galliani2015massively} proposed Gipuma, a massively parallel multi-view extension of PatchMatch MVS which leverages a red-black checkerboard pattern to perform a diffusion-like propagation scheme. This makes better use of the parallelism of GPUs to improve efficiency. Nonetheless, the major drawback is that each pixel's sampling points are preset, which in turn causes regional information to be ignored. Further, Xu and Tao ~\cite{xu2019multi} designed a method with Adaptive Checkerboard sampling and Multi-Hypothesis joint view selection (ACMH) to address this shortcoming. In allusion to depth estimation for low-textured areas, they further combined ACMH with multi-scale geometric consistency guidance to present ACMM. In addition, MARMVS ~\cite{xu2020marmvs} selects the optimal scale for each pixel to alleviate the ambiguity caused by regions with raw texture. Although the aforementioned approaches significantly enhance the performance of PatchMatch MVS method, 
how to recover the depth information of low-textured regions is always a challenging problem.

\begin{figure*}[htbp]
  \centering
  \includegraphics[width=14cm,page={1}]{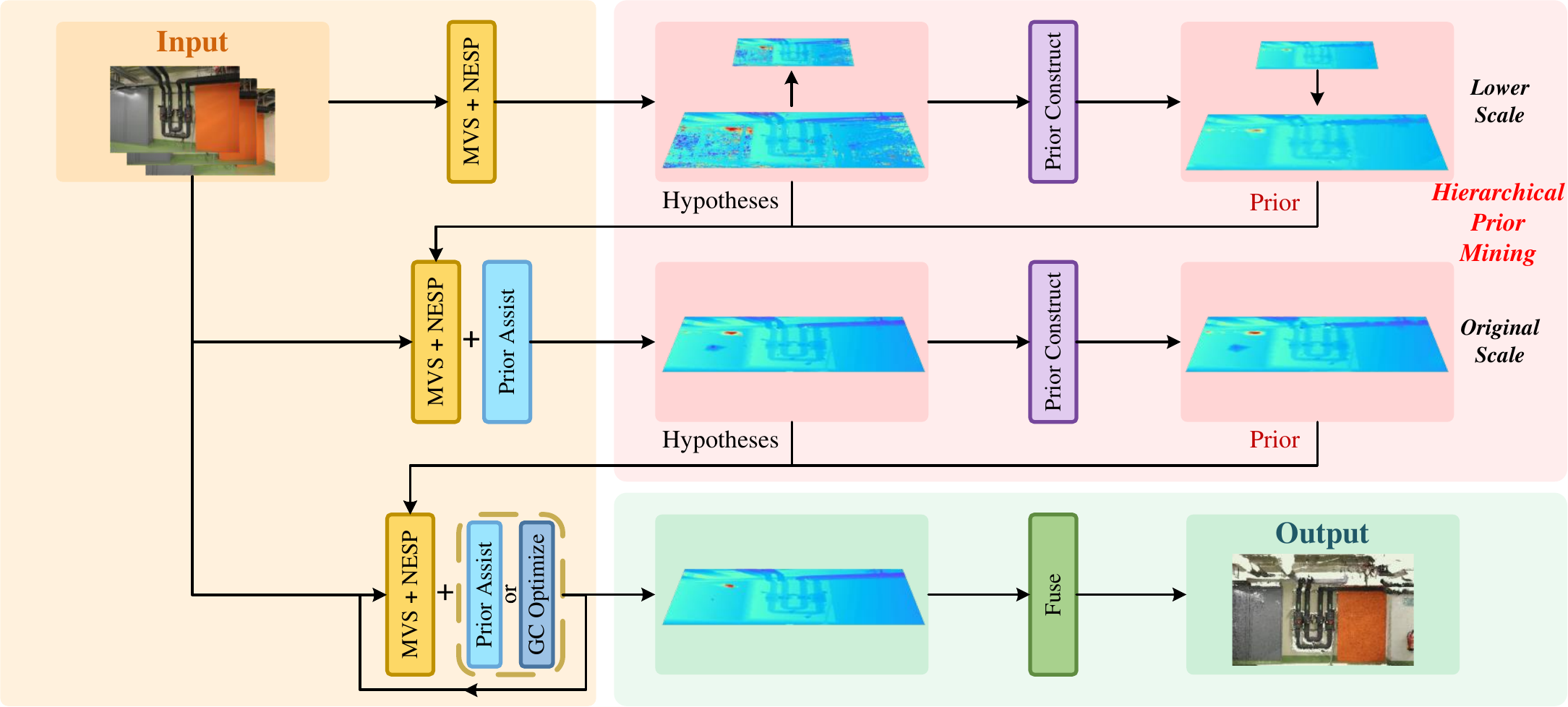}
  \caption{\textbf{An overview of HPM-MVS.} Starting with the input images, we first apply the basic MVS with NESP (Fig.~\ref{fig: Non-local Extensible Sampling Pattern}) to obtain the initial hypotheses. Then, we downsample them to the coarsest scale and generate the prior model. Third, we upsample the model to the current scale and leverage it to assist hypothesis prediction. After operating on different scales, we further use geometric consistency (GC) to optimize the results. Finally, the hypotheses will be fused to a point cloud.
  }
  \label{fig:pipeline}
\end{figure*}
{\bf \noindent Planar Prior Assistance.}  In order to deal with this long-standing ill-posed problem, the planar prior is considered. Romanoni \etal~\cite{romanoni2019tapa} proposed TAPA-MVS, they separated the image into two different scales of superpixels and utilized these as regional structures. The authors fitted a plane within the superpixels by using trustworthy 3D points and the RANSAC method, then leveraged the prior hypotheses to optimize the low-textured regions. 
ACMP~\cite{xu2020planar} constructs a planar prior depth map based on triangulations, and subsequently mosaics the prior hypotheses into the MVS process via a probabilistic graphical model. Recently, Xu \etal~\cite{xu2022multi} designed a  multi-scale geometric consistency guided and planar prior assisted Multi-View Stereo (ACMMP) by combining ACMM~\cite{xu2019multi} and ACMP~\cite{xu2020planar}. It fully exploits depth information at various scales and leverages prior assistance to efficiently target low-textured areas. Previous approaches only consider the prior information from the current image scale, we nonetheless believe that prior hypotheses should not be restricted to just one scale. To this aim, we embrace an HPM framework to discover more meaningful and comprehensive non-local prior knowledge.

\subsection{Learning-based Multi-View Stereo}
Deep learning has been widely used in 3D vision and has proven to be critical for MVS. MVSNet~\cite{yao2018mvsnet} introduces an end-to-end deep learning network that builds the 3D cost volumes from 2D image features. 
R-MVSNet~\cite{yao2019recurrent} regularizes the cost volume along the depth direction with the convolutional GRU to make reconstruction feasible. To further decrease memory and boost performance, CIDER~\cite{xu2020learning} builds a lightweight cost volume using an average group-wise correlation similarity metric. In recent years, the efficiency of learning-based MVS gains more attention. CVP-MVSNet~\cite{yang2020cost}, CasMVSNet~\cite{gu2020cascade} and UCS-Net~\cite{cheng2020deep} build a cascade cost volume by integrating the coarse-to-fine strategy. 
Furthermore, PatchMatchNet~\cite{wang2021patchmatchnet} embeds PatchMatch MVS into a deep learning framework to achieve high-quality reconstruction with low memory consumption.

Although the existing deep learning methods have achieved great success in MVS, they usually require a large quantity of training datasets, which may not be practical in real-world applications. In addition, the generalization ability still remains an issue for learning-based methods~\cite{keskar2016large}. As such, we still focus on traditional methods for MVS.

\section{Methodology}
In this section, we first briefly review the classical PatchMatch MVS. Then, we detail our proposed HPM-MVS. An overview of the method is illustrated in Fig.~\ref{fig:pipeline} (the pseudo-code is shown in the supplementary).

\subsection{Review of PatchMatch MVS}
PatchMatch MVS~\cite{galliani2015massively, schonberger2016pixelwise, xu2019multi} is a method for rapidly finding correspondences between image patches. To compute depth maps for input images, each image is selected in turn as a reference image and its corresponding source images are determined based on the Structure-from-Motion (SfM) results. 
In general, the process consists of four steps~\cite{xu2022multi}, i.e., random initialization, hypothesis propagation, multi-view cost evaluation and refinement. The last three steps are iterated until convergence, the details are as follows:

{\noindent 1) Random initialization} means randomly generating a plane hypothesis for each pixel in the reference image based on the results of SfM. A key observation behind is that a non-trivial fraction of a large field of random offset assignments is likely to be a good guess.

{\noindent 2) Hypothesis propagation} samples the hypotheses from neighboring pixels because of the high spatial coherence of neighbors in images. Moreover, a proper hypothesis can be propagated to a relatively large region to ensure the smoothness constraint of nearby pixels.

{\noindent 3) Multi-view matching cost evaluation} aims at robustly integrating matching cost from multiple views to select the best plane hypotheses.

{\noindent 4) Refinement} generates two additional hypotheses to enrich the diversity of the solution space. The one with the lowest cost will be selected as the refined estimation.

In this work, we concentrate on hypothesis propagation and multi-view matching cost evaluation, which are essential components of PatchMatch MVS methods, to capture non-local information for recovering better geometry in low-textured areas.

\subsection{Non-local Extensible Sampling Pattern}

Hypothesis propagation is a prerequisite for multi-view matching cost evaluation and refinement. Hence, choosing a reasonable sampling pattern will lead to accurate results. The pixels within a relatively large area can be represented by one of the pixels. So considering local information repeatedly is redundant. Our NESP is inspired by these two key insights to sample more robust hypotheses. 
Following the PatchMatch MVS pipeline introduced above, we design our basic MVS with NESP.

{\bf \noindent Random Initialization.} First, we randomly generate a hypothesis $\bm{\theta}_x  = [{d_x},\bm{n}_x]$ for each pixel $x$, where ${d_x}$ represents the depth information and $\bm{n}_x$ is a normal vector. Second, a matching cost is calculated from each source image via homography~\cite{hartley2003multiple}. Finally, the initial multi-view matching cost is obtained by averaging top-$k$ best costs.

\begin{figure}
  \centering
  \includegraphics[height=4cm,page={1}]{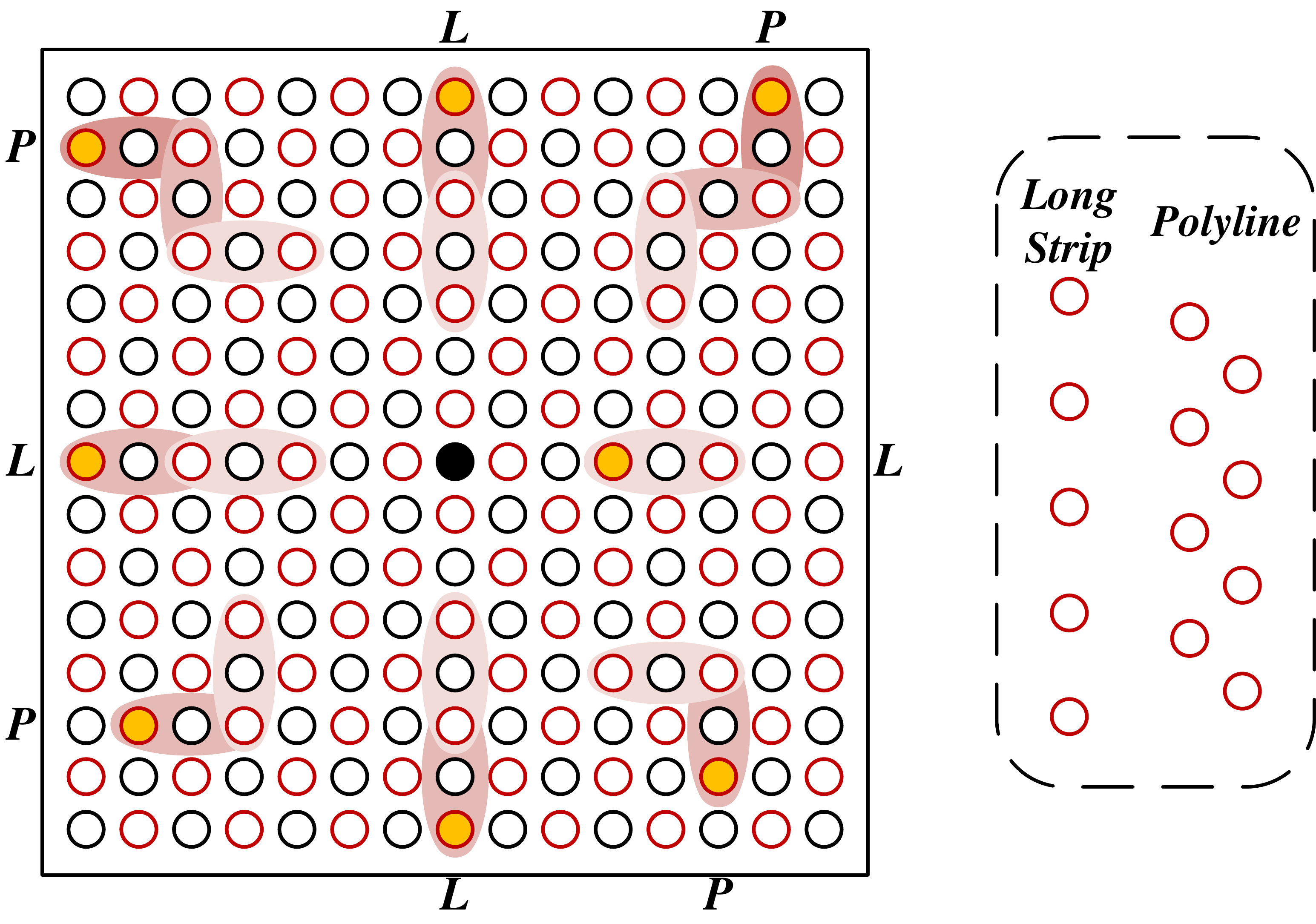}
  \caption{\textbf{Non-local Extensible Sampling Pattern.} The circles represent the pixels in the red-black checkerboard. The red areas represent the sampling regions; the lighter the color, the less adaptive extension is performed. The solid yellow circles show the sampled points.}
  \label{fig: Non-local Extensible Sampling Pattern}
\end{figure}

{\bf \noindent Hypothesis Propagation.} The diffusion-like propagation scheme can tremendously enhance efficiency while maintaining high-quality hypothesis propagation. Following~\cite{xu2019multi}, we first partition all pixels in the reference image into a red-black checkerboard pattern which uses hypotheses within eight regions as candidates to update. Second, we divide eight sampling areas into four long strip areas and four polyline areas; each polyline area starts with 8 samples, compared to each long strip area's initial 5 samples. Then, we consider two strategies to construct our NESP (Fig.~\ref{fig: Non-local Extensible Sampling Pattern}).

\begin{itemize}
\item \textbf{Extensible Strategy.} To achieve the purpose of extension, we first choose the best hypothesis in each area according to the multi-view matching costs from random initialization or the previous iteration (the initial costs in random initialization are used for the first propagation, while the other iterations use the costs computed in the previous iteration) and define the candidate hypothesis set as $\bm{\Theta} = \{ {\bm{\theta} _i}|i = 1 \cdot  \cdot  \cdot 8\}$. Second, we calculate their matching costs with respect to different source images and embed them into a cost matrix,
\begin{equation}
    \bm{M} = \left[ {\begin{array}{*{20}{c}}
    {{m_{1,1}}}&{{m_{1,2}}}& \cdots &{{m_{1,N - 1}}}\\
    {{m_{2,1}}}&{{m_{2,2}}}& \cdots &{{m_{2,N - 1}}}\\
    \vdots & \vdots & \ddots & \vdots \\
    {{m_{8,1}}}&{{m_{8,2}}}& \cdots &{{m_{8,N - 1}}}
    \end{array}} \right],
\end{equation}
where $N$ is the number of input images, ${m_{i,j}}$ represents the matching cost of the sampled point in the \emph{i}-th region scored by the \emph{j}-th view. Based on this matrix, a voting scheme is implemented in each line to determine
whether this region needs to be extended. Two thresholds are defined here: 1) A good matching cost threshold is,
\begin{equation}
    \tau ({t_{iter}},{t_{ext}}) = {\tau _{good}} \cdot {e^{ - \frac{{{{t_{iter}}^2} \cdot ({N_{ext}}  - {t_{ext}})}}{\alpha }}},
\end{equation}
where ${t_{iter}}$ and ${t_{ext}}$ represent the ${t_{iter}}$-th iteration of hypothesis propagation and the ${t_{ext}}$-th regional expansion, respectively; ${\tau _{good}}$ is the initial good matching cost threshold, $\alpha$ is a constant and ${N_{ext}}$ represents the maximum number of extensions. 2) A bad matching cost threshold is defined as ${\tau _{bad}}$. For a specific region, there should exist at least ${n_{good}}$ matching costs smaller than $\tau ({t_{iter}},{t_{ext}})$, and at most ${n_{bad}}$ matching costs meeting the condition: ${m_{i,j}} > {\tau _{bad}}$. When the above conditions are satisfied, the expansion of the current sampling area will be stopped. Otherwise, this area should extend by itself while each extension doubles the number of sampling points in the area. Notably, the set $\bm{\Theta}$ and the matrix $\bm{M}$ are renewed dynamically in each expansion using the rule described above.

\item \textbf{Non-local Strategy.} There are two key insights behind the non-local strategy: 1) The surrounding pixels of the central point share the same structured region information, and their hypotheses can be contained by Eq.~\ref{seven hypotheses} when in the process of refinement. 2) The misestimates in low-textured areas always appear as small isolated speckles in the depth map; this phenomenon indicates that these regions have fallen into local optimal solutions. Therefore, we use the non-local strategy (Fig.~\ref{fig: Non-local Extensible Sampling Pattern}) which abandons the sampling points within the radius $R$ of the center pixels. The motivation is to concentrate more on non-local areas instead of focusing on the local surrounding information.
\end{itemize}

In the end, we combine these two strategies to form NESP, which helps to collect more reasonable candidate hypotheses from non-local neighboring pixels.

{\bf \noindent Multi-View Matching Cost Evaluation.} In order to determine the best hypothesis from the above candidate hypothesis set, we follow the previous studies~\cite{zheng2014patchmatch, schonberger2016pixelwise, xu2019multi} to define the multi-view matching cost via photo consistency in our basic MVS with NESP,
\begin{equation}
{c_{photo}}(\bm{\theta} _i) = \frac{{\sum\nolimits_j {{w_j} \cdot {m_{i,j}}} }}{{\sum\nolimits_j {{w_j}} }},
\label{matching_cost}
\end{equation}
where $w_j$ is the weight of the $j$-th view, which is computed by the view selection strategy in \cite{xu2019multi}. The hypothesis with the least matching cost will be chosen as the current best estimate for the center pixel.

{\bf \noindent Refinement.} We generate two new hypotheses in refinement. One hypothesis is obtained by perturbing the current hypothesis $[d_x^{p},\bm{n}_x^{p}]$, while the other is randomly generated $[d_x^{r},\bm{n}_x^{r}]$. Subsequently, the two new hypotheses are randomly arranged and combined with the current hypothesis to form an ensemble of seven hypotheses,
\begin{equation}
    \begin{array}{l}
    \{ [{d_x},\bm{n}_x],[{d_x},\bm{n}_x^{p}],[{d_x},\bm{n}_x^{r}],[d_x^{r},\bm{n}_x],\\
    \quad\quad[d_x^{r},\bm{n}_x^{r}],[d_x^{p},\bm{n}_x],[d_x^{p},\bm{n}_x^{p}]\}. 
\end{array}
\label{seven hypotheses}
\end{equation}
The final estimation for pixel \emph{x} will be the hypothesis with the minimum cost.

\subsection{Planar Prior Construction}
The local image patches of low-textured areas are highly similar, and the photometric consistency, as a metric function of MVS, always leads to incorrect estimates due to its defects. Inspired by ~\cite{xu2020planar}, we construct a better planar prior model to assist multi-view matching cost evaluation. After executing the basic MVS, a hypothesis with its final cost for each pixel can be obtained. A lower cost represents a more precise estimate, therefore we collect the credible correspondences into a set $\bm{I}_{cred}$,
\begin{equation}
    {\bm{I}_{cred}} = \{ {\bm{\theta}_x}|{c_{photo}}({\bm{\theta} _x}) < {\tau _{cred}}\},
\end{equation}
where ${{\tau _{cred}}}$ represents the credible threshold. Then, the Delaunay triangulation is applied to generate triangular surfaces of various sizes. Although these triangles cover a large area, there are still some marginal areas in the image that remain unaffected. This suggests that the local information cannot satisfy the requirements of developing a planar prior model in these parts of the regions. To address this issue, we propose a method based on KNN to construct the planar prior model via non-local credible hypotheses. First, we set up a KD-tree~\cite{zhou2008real} for the reliable points in $\bm{I}_{cred}$. Second, the top-$K$ nearest non-local neighbors are searched for pixels in the regions where prior construction is challenging. In addition, we use Heron's formula to avoid the case of co-linearity of proximity points. Then, each pixel will have three reliable neighboring points. We project them into the coordinate of the reference camera and set up a matrix $\bm{A}$ to get the plane parameters $\bm{z}_{opt}$,
\begin{equation}
    {\bm{z}_{opt}} = \mathop {\arg \min }\limits_{{\bm{z}^*}:||{\bm{z}^*}|| = 1} ||{\bm{A}} \cdot {\bm{z}^*}||,
\end{equation}
where $\bm{z}^*$ is a unit-length variable. To further enhance the validity of the prior model, the ultimate prior knowledge for each image is filtered by its depth range. At last, we consider Planar Prior Assistance~\cite{xu2020planar} which can better leverage the planar prior model to assist multi-view matching cost evaluation in low-textured areas.
\subsection{Hierarchical Prior Mining}

\begin{figure}[t]
	\centering
	\begin{subfigure}{0.325\linewidth}
		\centering
		\includegraphics[width=1\linewidth]{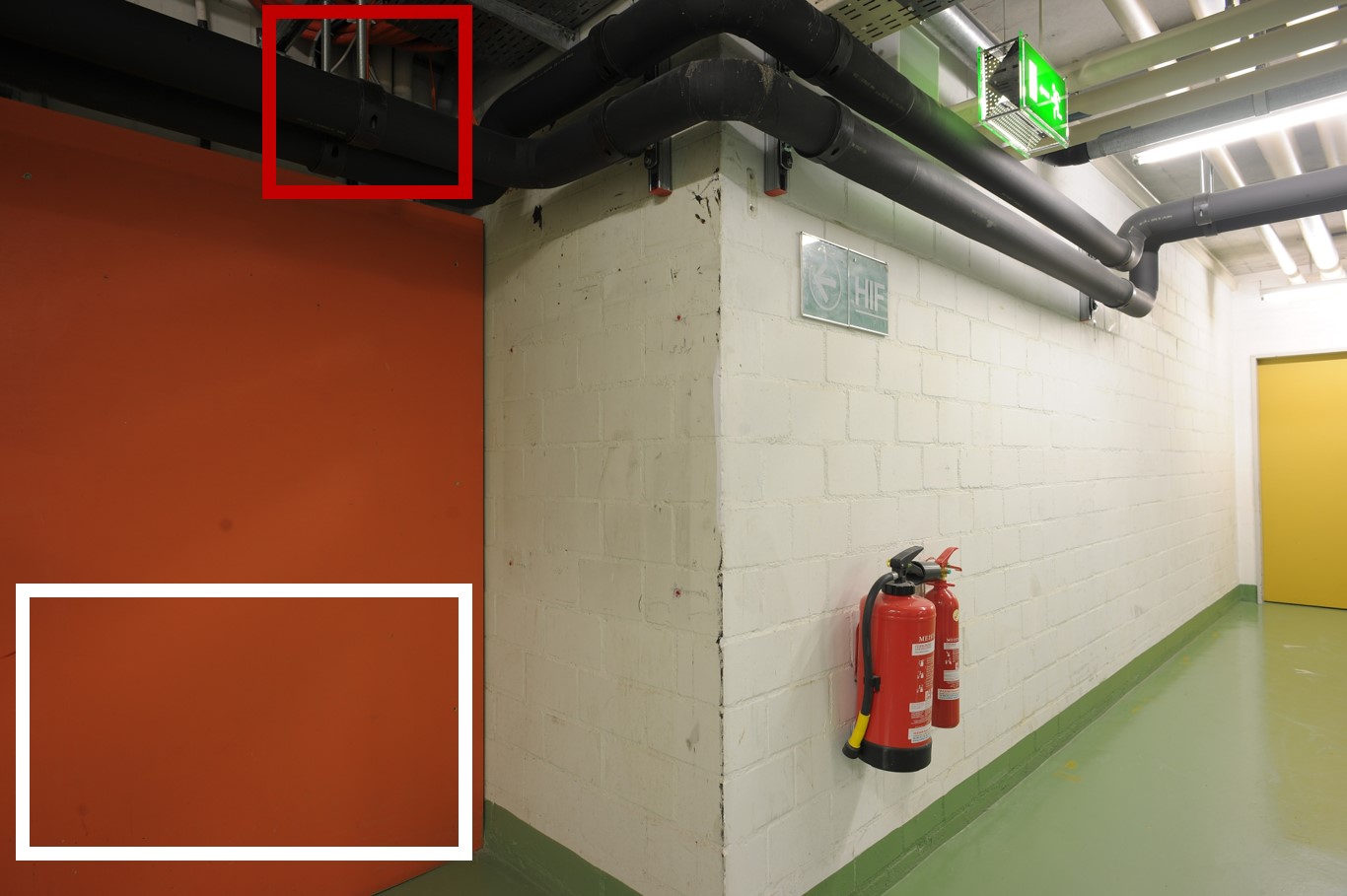}
		\caption{}
		\label{prior-RGB}
	\end{subfigure}
        \centering
	\begin{subfigure}{0.325\linewidth}
		\centering
		\includegraphics[width=1\linewidth]{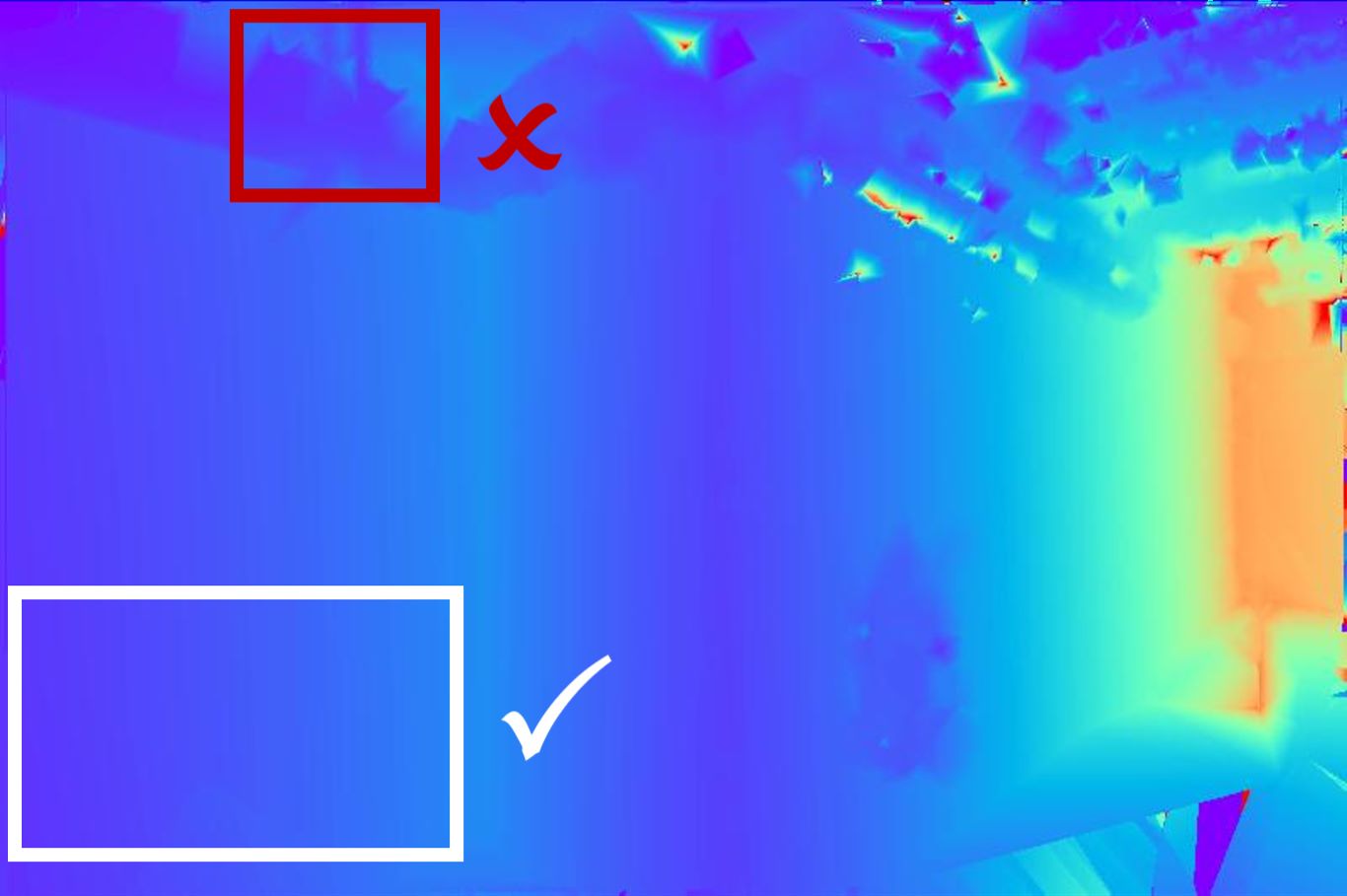}
		\caption{}
		\label{prior-low}
	\end{subfigure}
	\centering
	\begin{subfigure}{0.325\linewidth}
		\centering
		\includegraphics[width=1\linewidth]{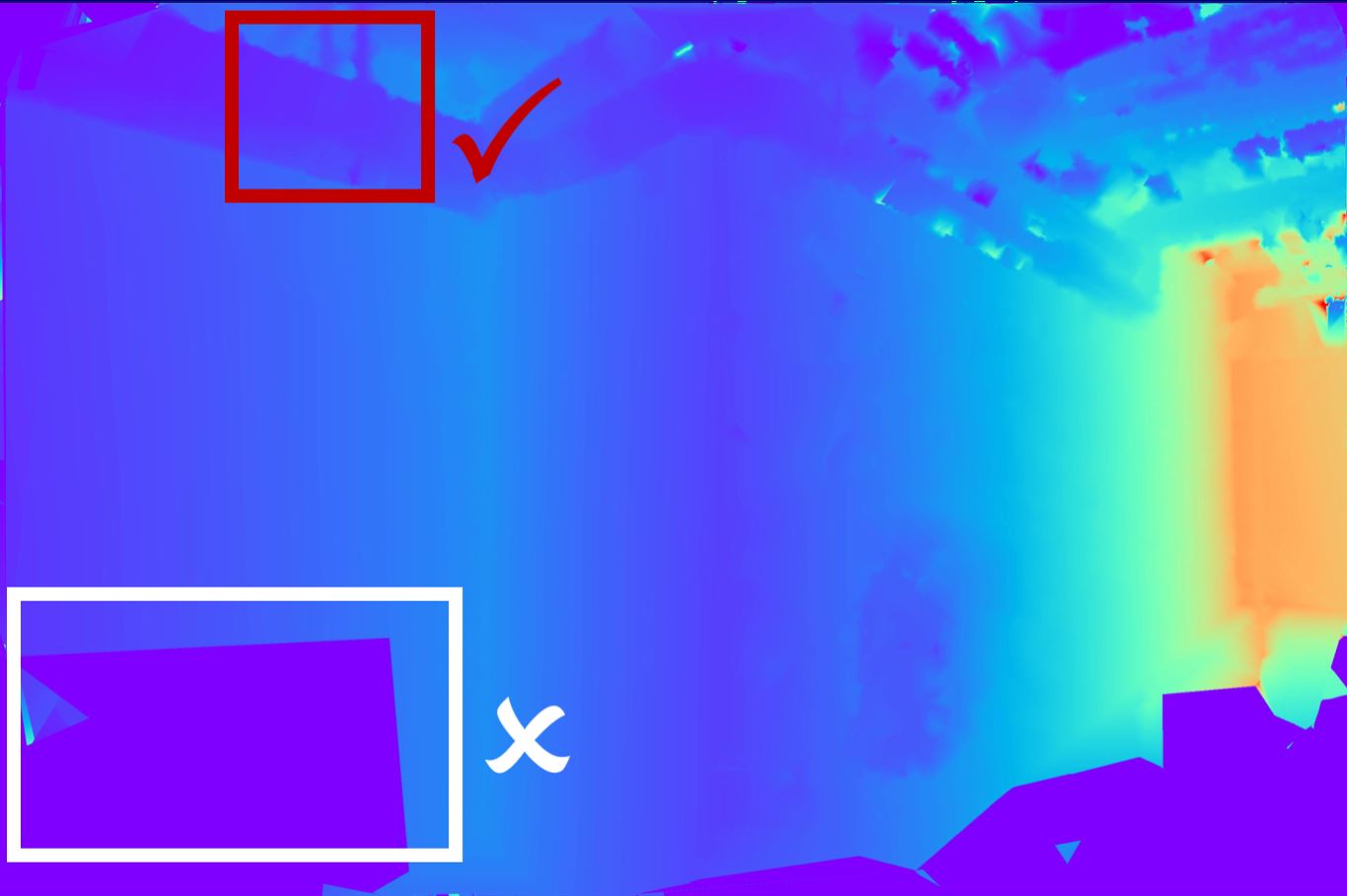}
		\caption{}
		\label{prior-high}
	\end{subfigure}
	\centering
  \begin{subfigure}{0.325\linewidth}
		\centering
		\includegraphics[width=1\linewidth]{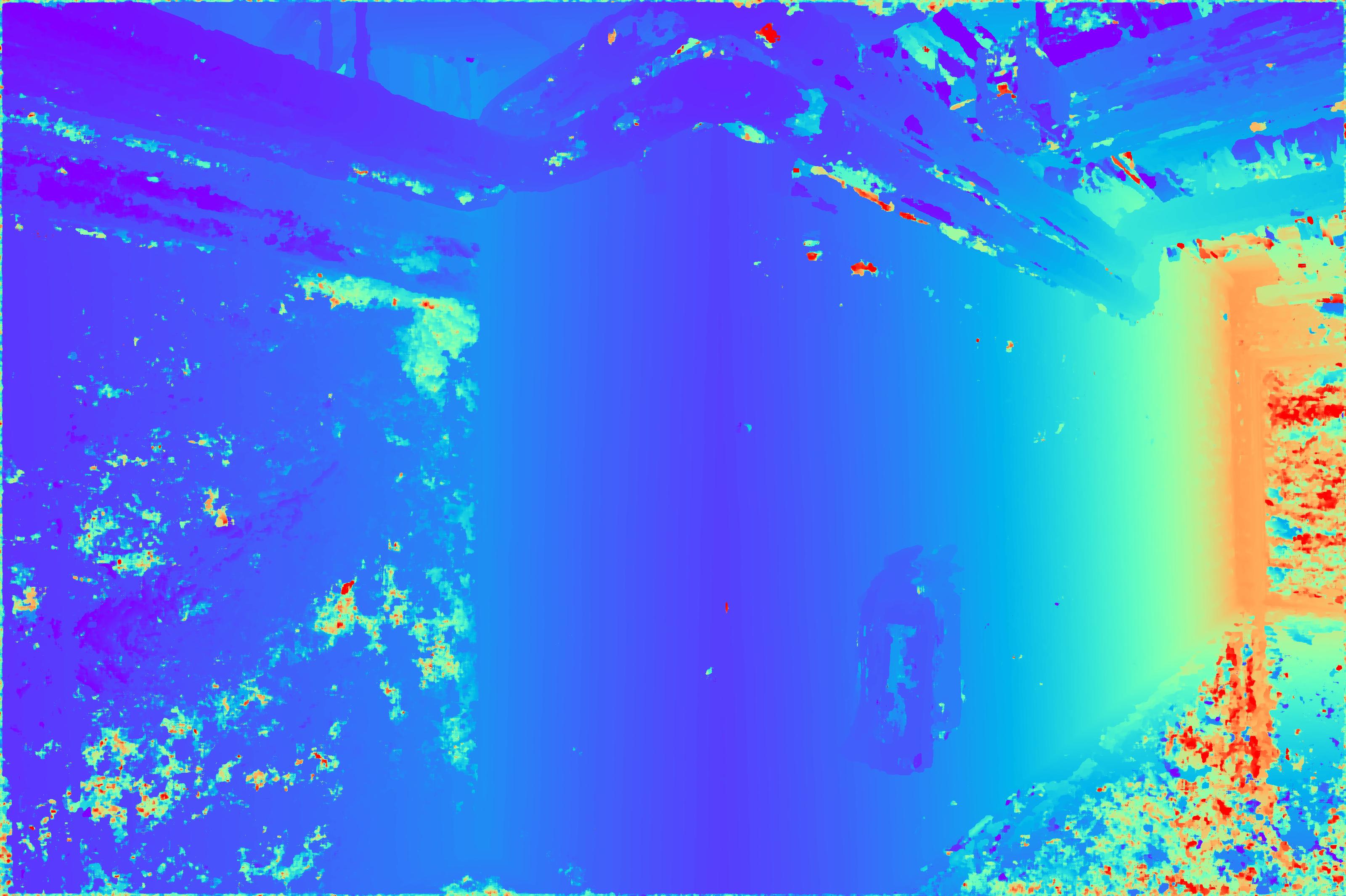}
		\caption{}
		\label{basic}
	\end{subfigure}
	\centering
 \begin{subfigure}{0.325\linewidth}
		\centering
		\includegraphics[width=1\linewidth]{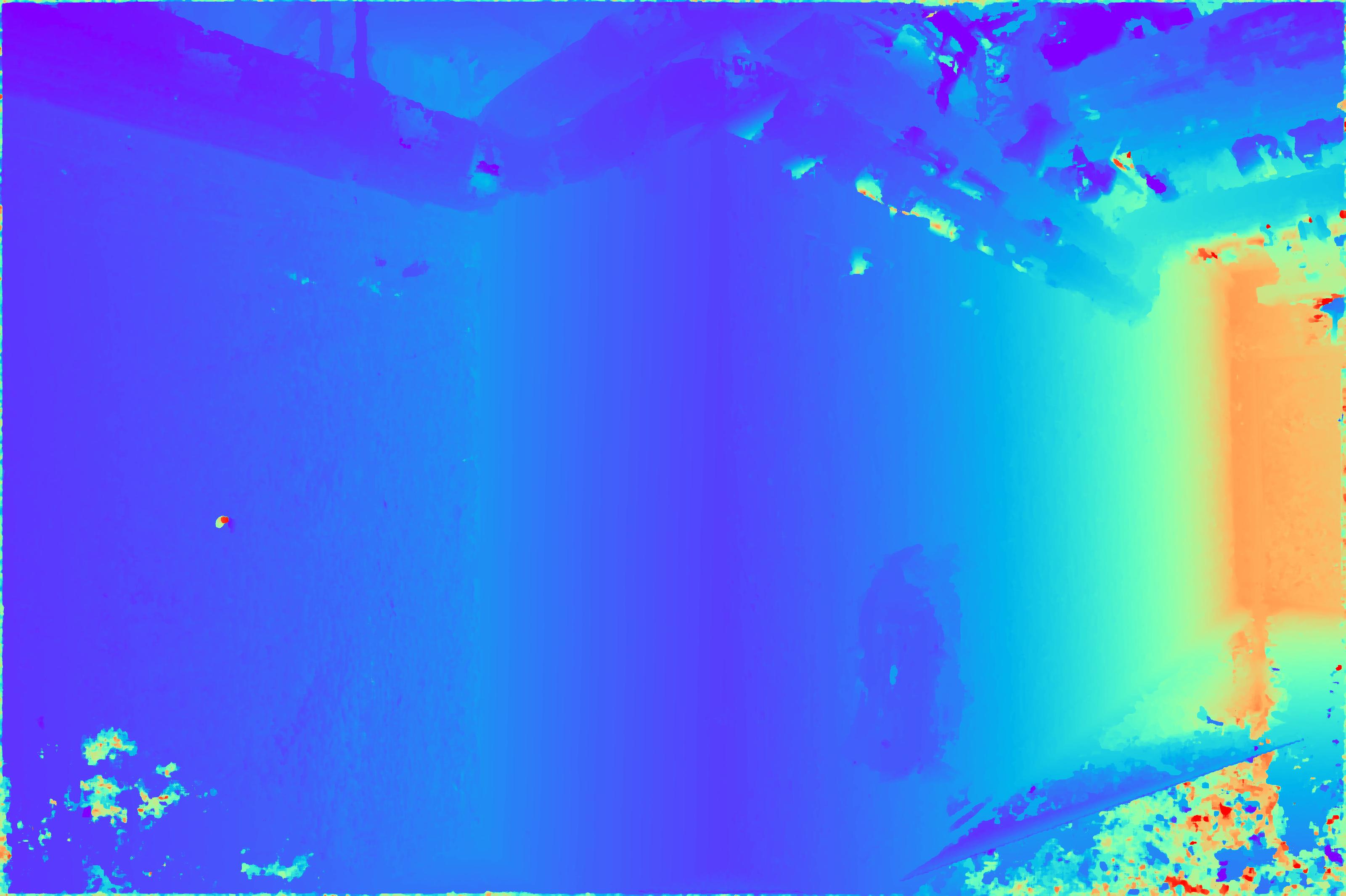}
		\caption{}
		\label{basic}
	\end{subfigure}
	\centering
 \begin{subfigure}{0.325\linewidth}
		\centering
		\includegraphics[width=1\linewidth]{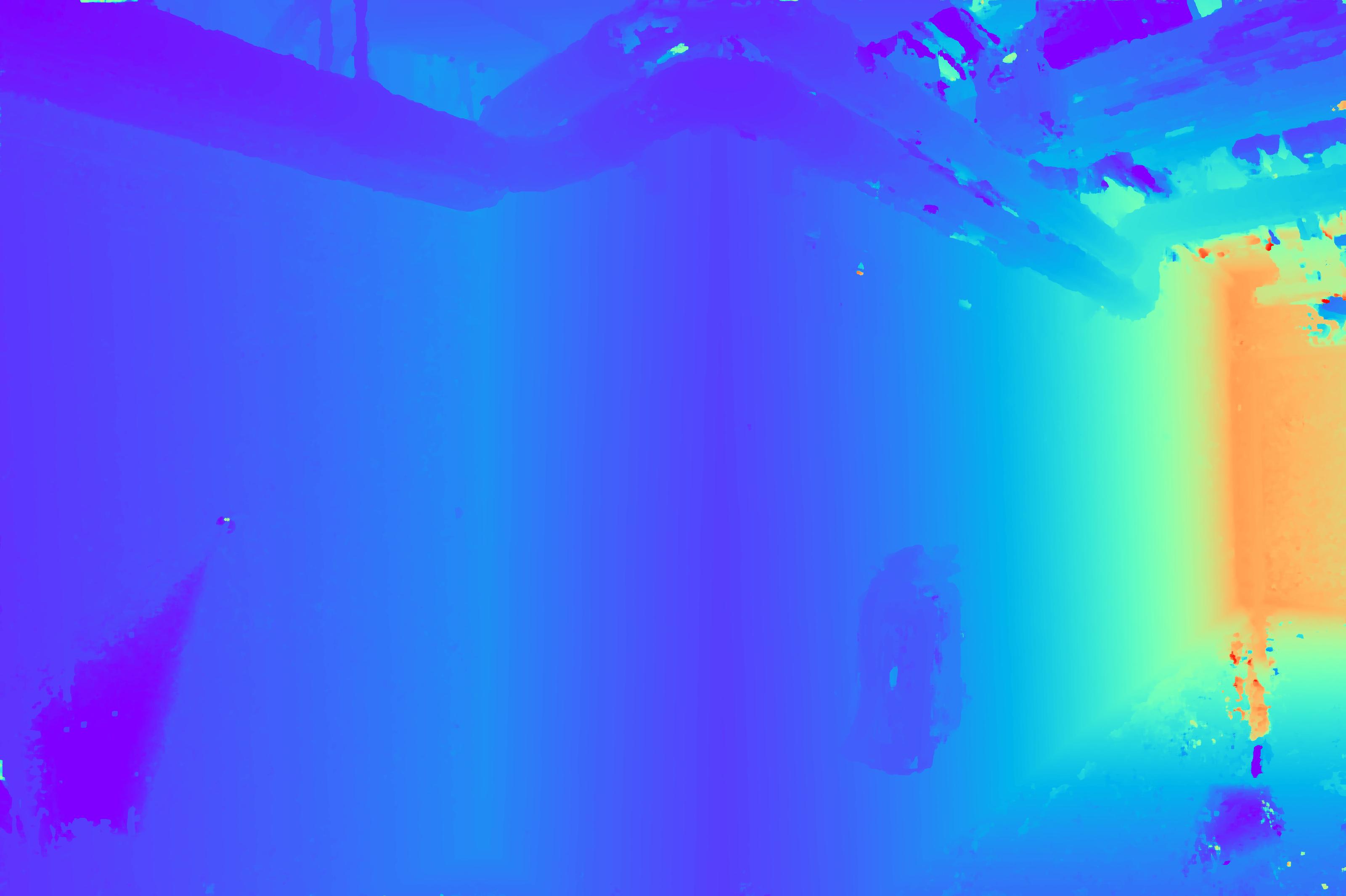}
		\caption{}
		\label{basic}
	\end{subfigure}
	\centering
	\caption{Planar prior models at different scales and resulting depth maps. (a) The color image (the low-textured area is in the white box while pixels in the red box are details); (b) the planar prior model constructed at the low-resolution scale; (c) the planar prior model constructed at the high-resolution scale (the dark blue areas show that there are anomalies and have already been filtered out); (d) the depth map obtained without prior models; (e) the depth map based on the prior model constructed at the low-resolution scale; (f) the depth map based on 
    prior models at different scales.}
	\label{prior}
\end{figure}

The multi-scale structure is employed in some PatchMatch MVS methods to infer the depth information of low-textured areas. Previous approaches~\cite{liao2019pyramid,xu2019multi,wang2020mesh} usually construct image pyramids to fuse depth estimation at different scales. In addition, several methods~\cite{xu2020planar,romanoni2019tapa} focus on constructing planar priors models at a single scale to help the depth estimation. In our work, we demonstrate that constructing planar prior models in a multi-scale way will further facilitate depth estimation.

As depicted in Fig.~\ref{prior}(b), when the planar prior model is built in low-textured regions, the performance at the low-resolution scale is obviously more advanced than the high-resolution scale. Because the information is condensed after downsampling, resulting in a larger receptive field for each pixel, and a wider range of non-local information can be excavated than the original scale. However, only using the prior from the coarsest scale may lead ambiguity in detail areas. We find that the aforementioned problem of prior construction is expertly managed at the high-resolution scale (Fig.~\ref{prior}(c)). 

Therefore, the depth perception advantages of planar prior models at different scales are complementary. Inspired by this key observation, we design an HPM framework to construct prior models at different scales, which assists the depth estimation of low-textured regions without sacrificing details. The depth map can be continually optimized via the HPM framework (Fig.~\ref{prior}(d)-(f)).

Excavating hierarchical prior is essentially an iterative coarse-to-fine optimization process (Fig.~\ref{fig:pipeline}). After employing the basic MVS method with NESP, we first downsample the obtained hypotheses (\ie depth and normal) to the coarsest scale and build a planar prior model. Second, with the joint bilateral upsampler~\cite{kopf2007joint}, we propagate the planar prior to the original scale. Third, the planar prior model is embedded into the PatchMatch MVS method with Planar Prior Assistance~\cite{xu2022multi} to approximate the depth in low-textured areas better. Finally, we downsample the newly generated hypotheses to the medium scale again, and repeat the same operation until iterating up to the original scale. For images with high resolutions, more medium scales can be considered. To further optimize the results, we apply geometric consistency as~\cite{schonberger2016pixelwise, xu2019multi} do. In this way, the final depth estimation can achieve a better trade-off between details and low-textured areas.

Moreover, to save computing resources and improve time efficiency, we propose a fast version of the HPM framework. This version only uses the planar prior model constructed at the coarsest scale to assist MVS. The rationale behind this is that the prior model obtained at the coarsest scale can effectively target the low-textured region and exactly compensate for the deficiencies of the basic MVS method in that region. Both HPM and its fast version will be experimentally analyzed.

\begin{figure*}[h]
        \begin{subfigure}{0.16\linewidth}
		\centering
		\includegraphics[width=1\linewidth]{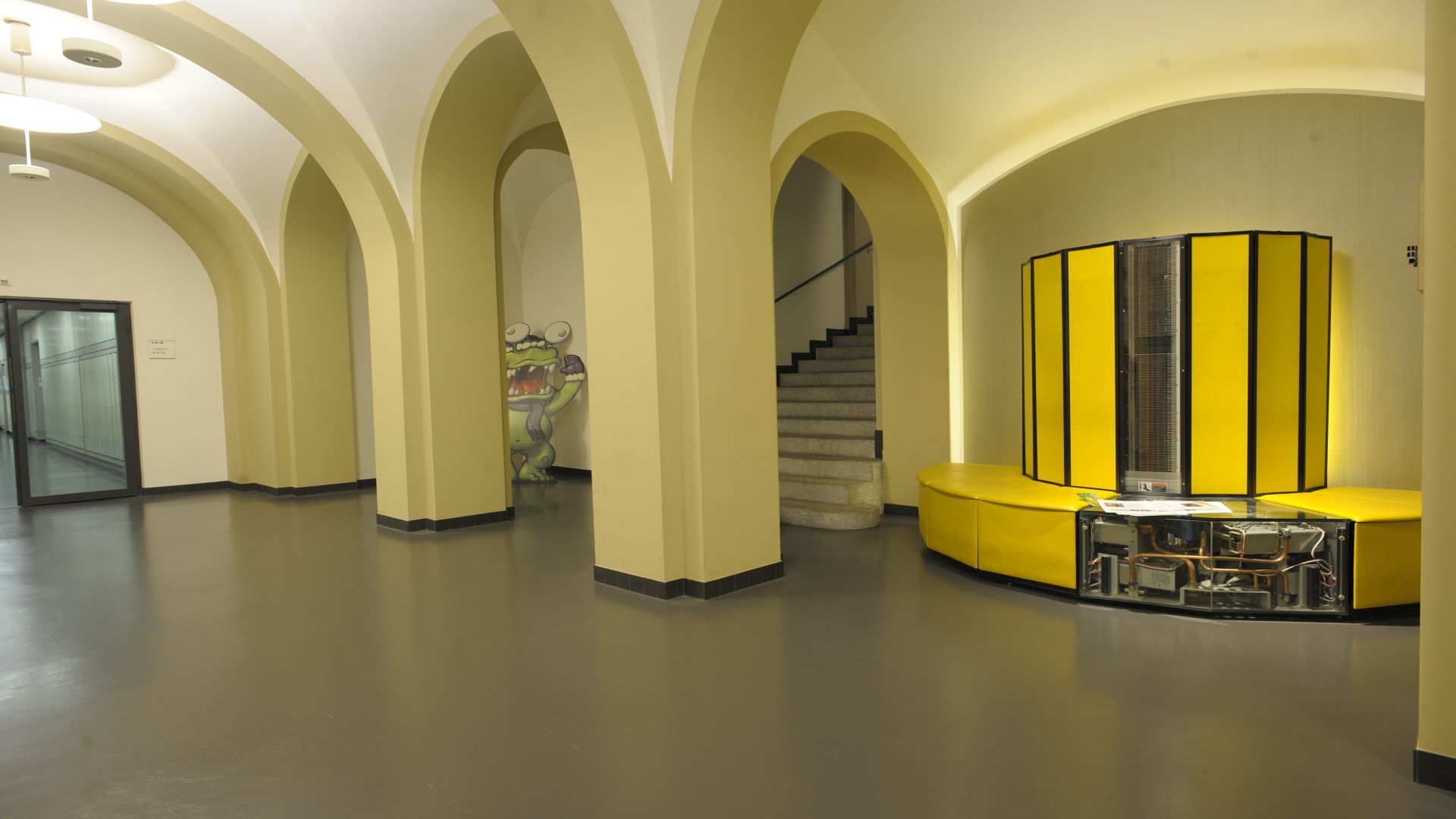}
	\end{subfigure}
	\begin{subfigure}{0.16\linewidth}
		\centering
		\includegraphics[width=1\linewidth]{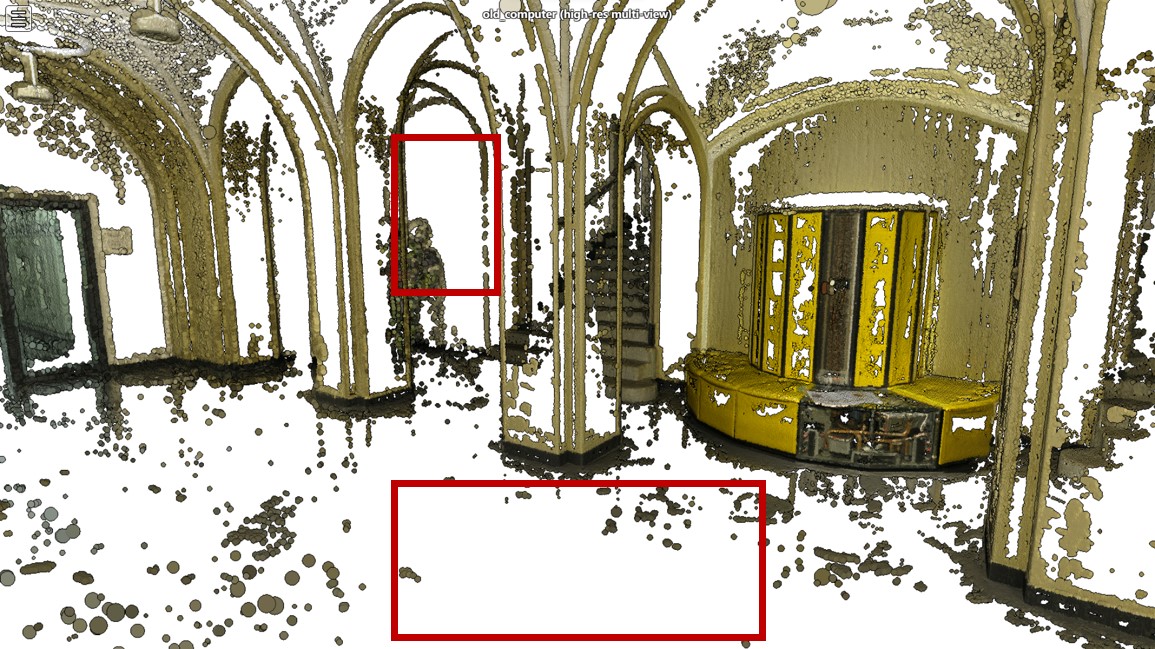}
	\end{subfigure}
        \begin{subfigure}{0.16\linewidth}
		\centering
		\includegraphics[width=1\linewidth]{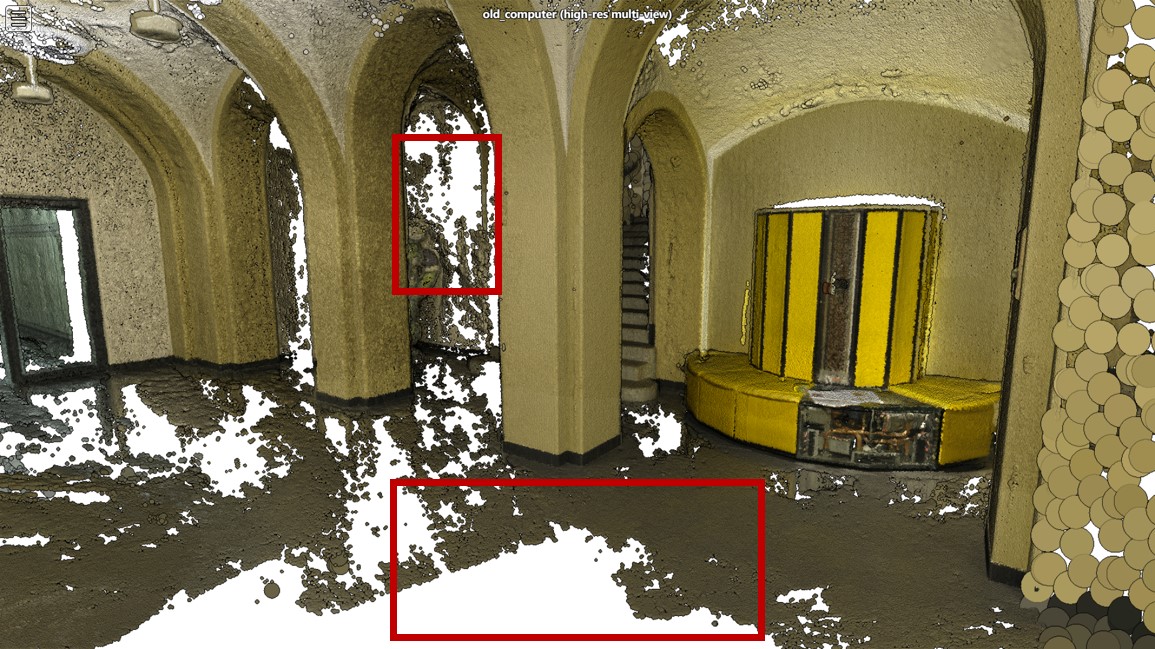}
	\end{subfigure}
        \begin{subfigure}{0.16\linewidth}
		\includegraphics[width=1\linewidth]{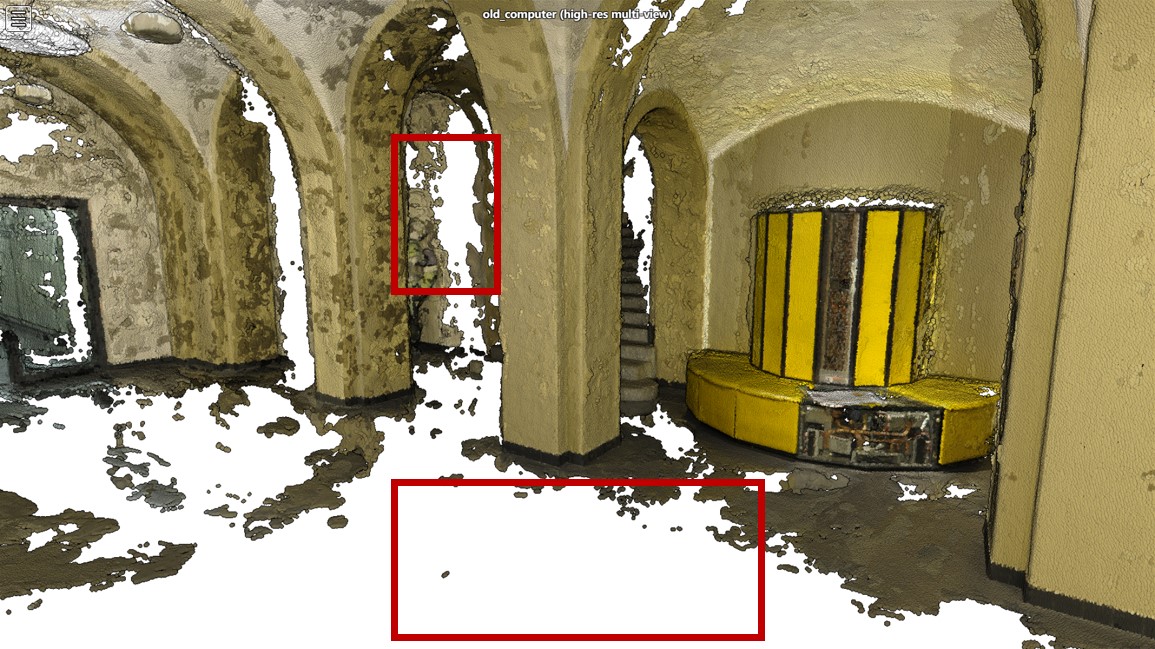}
	\end{subfigure}
        \begin{subfigure}{0.16\linewidth}
		\includegraphics[width=1\linewidth]{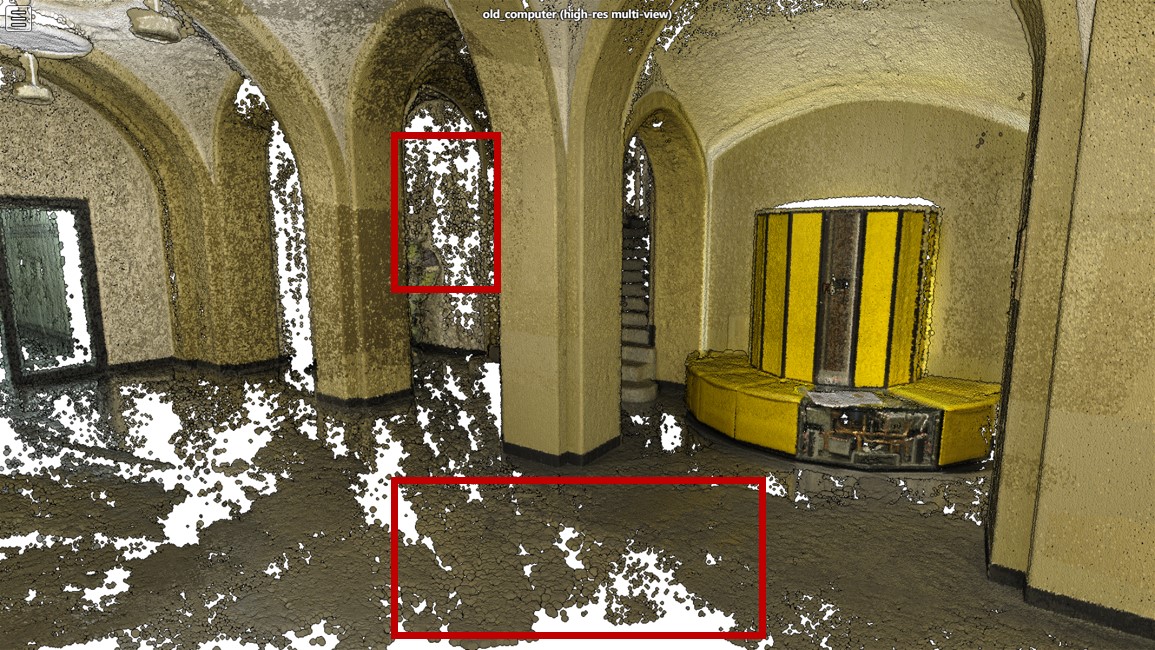}
	\end{subfigure}
        \begin{subfigure}{0.16\linewidth}
		\includegraphics[width=1\linewidth]{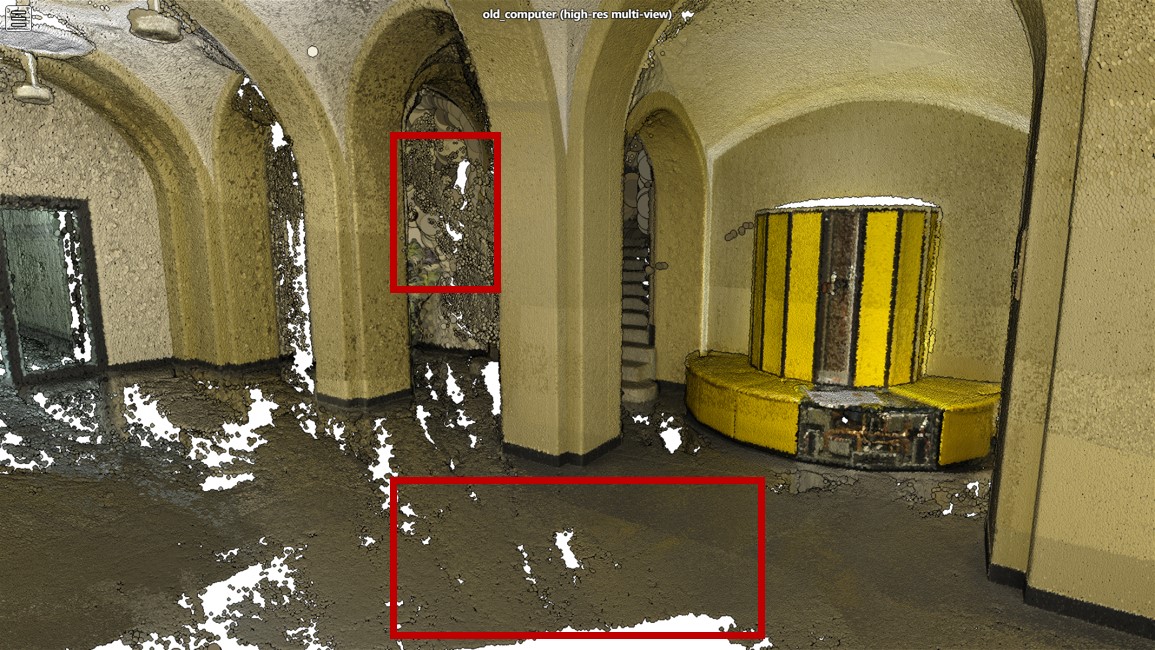}
	\end{subfigure}
 \centering
        \begin{subfigure}{0.16\linewidth}
		\centering
		\includegraphics[width=1\linewidth]{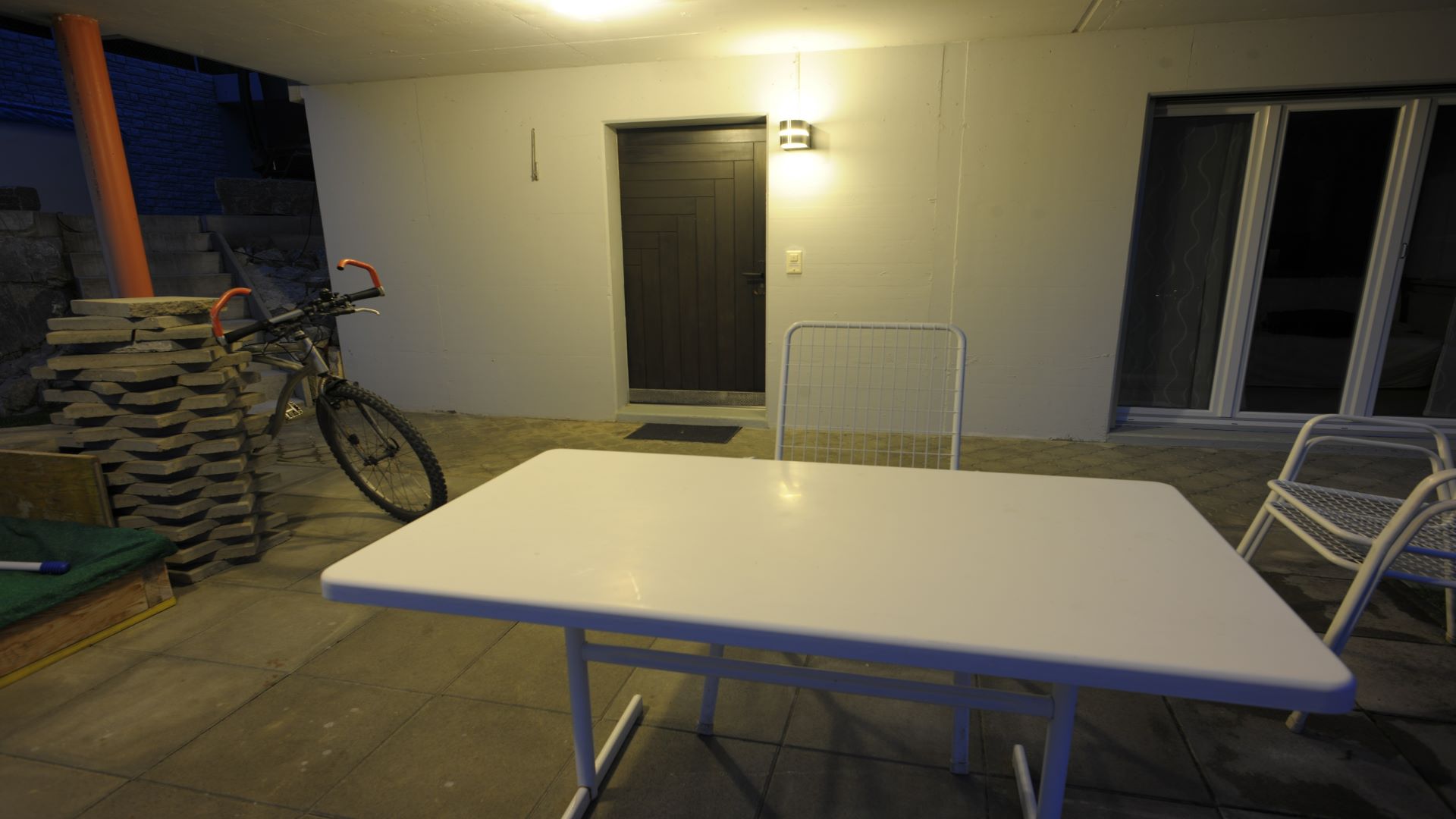}
  \caption{Images}
	\end{subfigure}
	\begin{subfigure}{0.16\linewidth}
		\centering
		\includegraphics[width=1\linewidth]{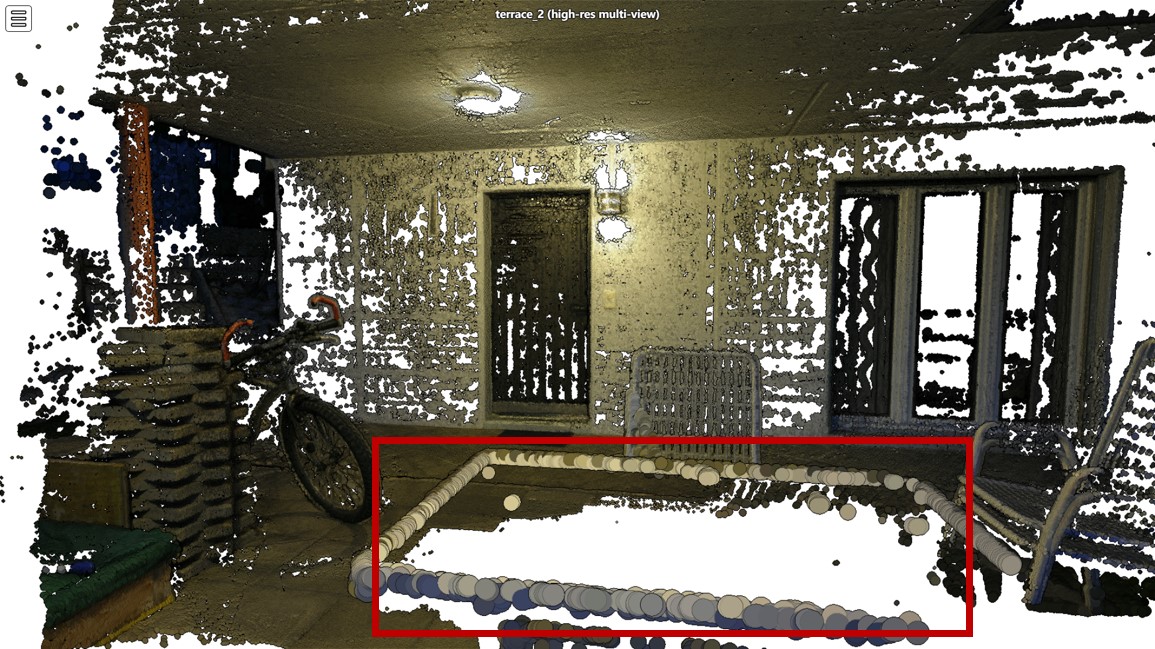}
  \caption{COLMAP~\cite{schonberger2016pixelwise}}
	\end{subfigure}
        \begin{subfigure}{0.16\linewidth}
		\centering
		\includegraphics[width=1\linewidth]{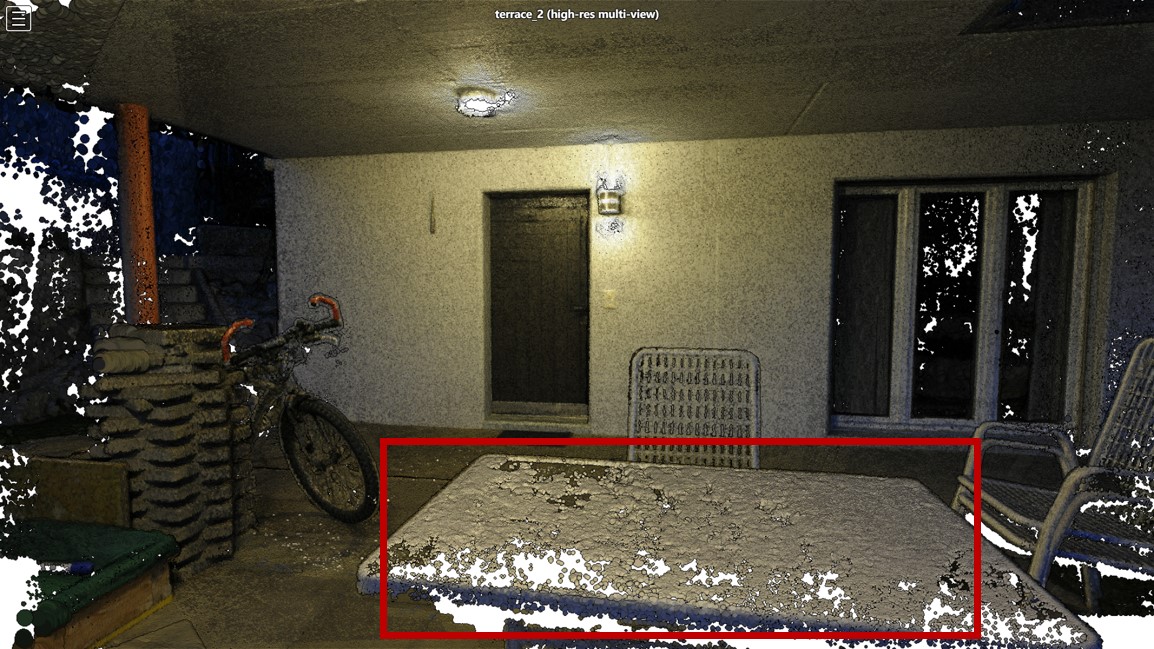}
  \caption{ACMMP~\cite{xu2022multi}}
	\end{subfigure}
        \begin{subfigure}{0.16\linewidth}
		\includegraphics[width=1\linewidth]{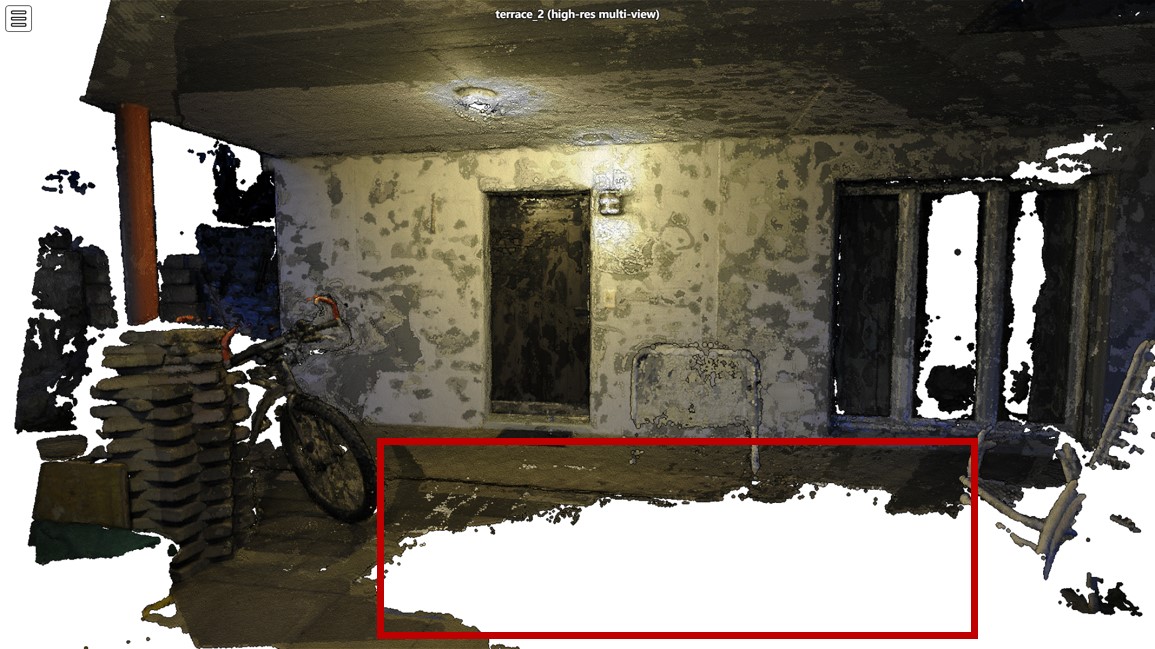}
  \caption{IterMVS~\cite{wang2022itermvs}}
	\end{subfigure}
        \begin{subfigure}{0.16\linewidth}
		\includegraphics[width=1\linewidth]{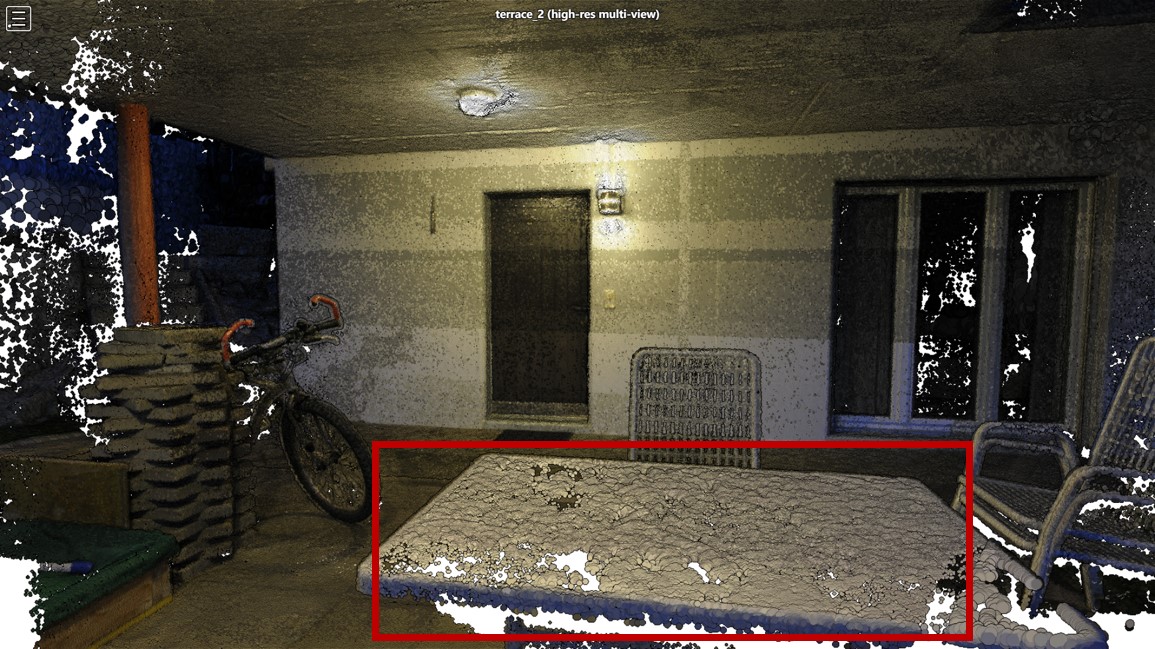}
  \caption{HPM-MVS$_{{\rm{fast}}}$}
	\end{subfigure}
        \begin{subfigure}{0.16\linewidth}
		\includegraphics[width=1\linewidth]{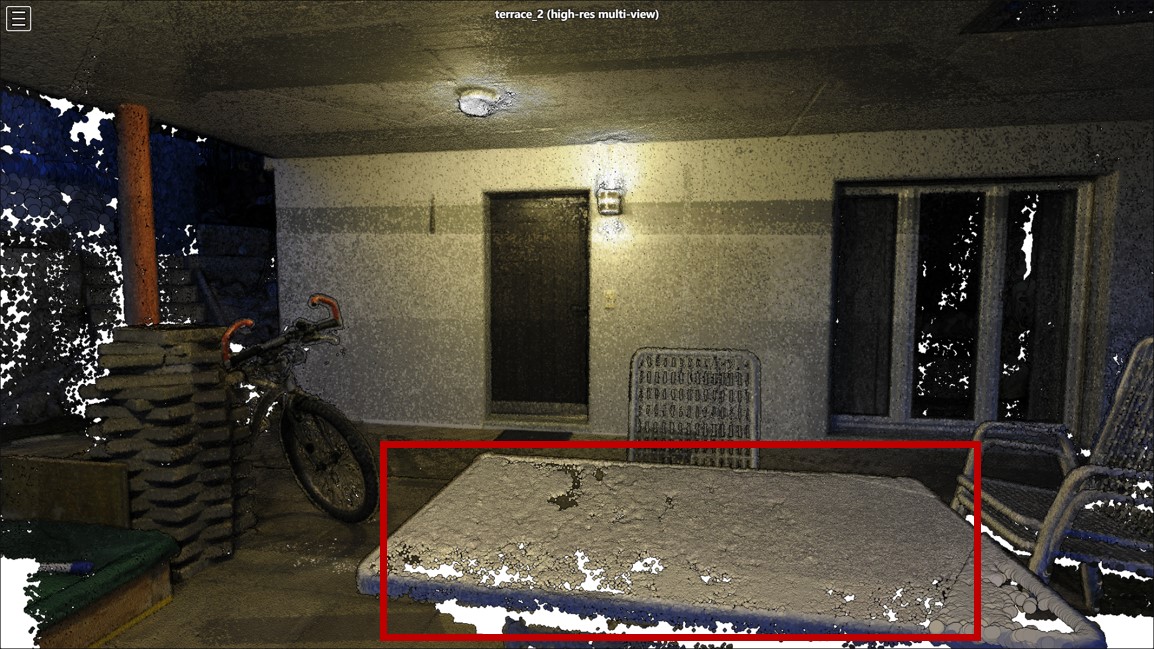}
  \caption{HPM-MVS}
	\end{subfigure}
 \centering
	\caption{Qualitative point cloud comparisons between different MVS methods on several high-resolution multi-view test scans (old computer and terrace 2) of ETH3D benchmark~\cite{schops2017multi}. Some challenging areas are shown in red boxes.}
	\label{visualize}
\centering
\end{figure*}

\section{Experiments}
We evaluate our method on two challenging MVS datasets, ETH3D benchmark~\cite{schops2017multi} and Tanks \& Temples datasets~\cite{knapitsch2017tanks}, from two perspectives, \ie, benchmark performance and analysis experiments.
\subsection{Experimental Setup}
ETH3D high-resolution multi-view benchmark~\cite{schops2017multi} consists of training datasets and test datasets, which contain indoor and outdoor images at a resolution of 6048 $\times$ 4032\footnote{We downsample the undistorted images to 3200 $\times$ 2130 as in ~\cite{schonberger2016pixelwise}.}. 
Tanks \& Temples datasets are divided into Intermediate and Advanced datasets based on the complexity of the scene. However, these datasets do not provide official camera poses. In that case, we use COLMAP~\cite{schonberger2016pixelwise} to estimate camera parameters.

All of our experiments are executed on a computer with an Intel i5-12600 CPU and an RTX 3070 GPU. In the basic MVS with NESP, \{$k$, $N_{ext}$, ${\tau _{{\rm{good}}}}$, ${\tau _{bad}}$, $\alpha $, ${n_{good}}$, ${n_{bad}}$, $R$\} = \{4, 3, 0.8, 1.2, 90, 1, 2, 4\}. In the process of planar prior construction, \{${\tau _{cred}}$, $K$\} = \{0.1, 6\}. 

\subsection{Benchmark Performance}

\begin{table}[]
\caption{Point cloud evaluation on ETH3D benchmark~\cite{schops2017multi}. We show accuracy / completeness / ${F_1}$ score (in $\%$) at different thresholds ($2cm$ and $10cm$). The best results are marked in \textbf{bold} while the second-best results are \underline{underlined}.}
\centering
    \resizebox{\linewidth}{!}{
\begin{tabular}{c|c|c|c}

\hline
                         & Method           & $2cm$                   & $10cm$                  \\ \hline
\multirow{12}{*}{Train.} & \textit{i) Traditional}   &                       &                       \\
                         & COLMAP~\cite{schonberger2016pixelwise}           & \textbf{91.85} / 55.13 / 67.66 & \textbf{98.75} / 79.47 / 87.61 \\
                         & PCF-MVS~\cite{kuhn2019plane}          & 84.11 / 75.73 / 79.42 & 95.98 / 90.42 / 92.98 \\
                         & ACMM~\cite{xu2019multi}             & 90.67 / 70.42 / 78.86 & 98.12 / 86.40 / 91.70 \\
                         & ACMP~\cite{xu2020planar}             & 90.12 / 72.15 / 79.79 & 97.97 / 87.15 / 92.03 \\
                         & ACMMP~\cite{xu2022multi}            & 91.03 / \underline{77.27} / \underline{83.42} & 97.96 / \underline{93.19} / \underline{95.46} \\ \cline{2-4} 
                         & \textit{ii) Learning} &                       &                       \\
                         & PatchMatchNet~\cite{wang2021patchmatchnet}    & 64.81 / 65.43 / 64.21 & 89.98 / 83.28 / 85.70 \\
                         & IterMVS~\cite{wang2022itermvs}          & 73.62 / 61.87 / 66.36 & 94.48 / 78.85 / 85.25 \\
                         & MVSTER~\cite{wang2022mvster}           & 68.08 / 76.92 / 72.06 & 91.97 / 91.91 / 91.73 \\ \cline{2-4} 
                         & HPM-MVS$_{{\rm{fast}}}$    & \underline{91.17} / 73.20 / 80.86 & \underline{98.23} / 92.41 / 94.97 \\
                         & HPM-MVS          & 90.66 / \textbf{79.50} / \textbf{84.58} & 97.97 / \textbf{95.59} / \textbf{96.22} \\ \hline
\multirow{12}{*}{Test}   & \textit{i) Traditional}   &                       &                       \\
                         & COLMAP~\cite{schonberger2016pixelwise}           & 91.97 / 62.98 / 73.01 & \underline{98.25} / 84.54 / 90.40 \\
                         & PCF-MVS~\cite{kuhn2019plane}          & 82.15 / 79.29 / 80.38 & 92.12 / 91.26 / 91.56 \\
                         & ACMM~\cite{xu2019multi}             & 90.65 / 74.34 / 80.78 & 98.05 / 88.77 / 92.96 \\
                         & ACMP~\cite{xu2020planar}             & 90.45 / 75.58 / 81.51 & 97.47 / 88.71 / 92.62 \\
                         & ACMMP~\cite{xu2022multi}            & 91.91 / 82.10 / \underline{85.89} & 98.05 / 94.67 / 96.27 \\ \cline{2-4} 
                         & \textit{ii) Learning} &                       &                       \\
                         & PatchMatchNet~\cite{wang2021patchmatchnet}    & 79.71 / 77.46 / 73.12 & 91.98 / 92.05 / 91.91 \\
                         & IterMVS~\cite{wang2022itermvs}          & 76.91 / 72.65 / 74.29 & 95.42 / 85.81 / 90.15 \\
                         & MVSTER~\cite{wang2022mvster}           & 77.09 / \underline{82.47} / 79.01 & 94.21 / 92.71 / 92.30 \\ \cline{2-4} 
                         & HPM-MVS$_{{\rm{fast}}}$    & \textbf{92.50} / 80.25 / 85.35 & \textbf{98.32} / \underline{94.89} / \underline{96.51} \\
                         & HPM-MVS          & \underline{92.13} / \textbf{83.25} / \textbf{87.11} & 98.11 / \textbf{95.41} / \textbf{96.69} \\ \hline

\end{tabular}
}
\label{tab:ETH3D_table}
\end{table}

{\bf \noindent{Evaluation on ETH3D Benchmark.}} Both traditional methods~\cite{schonberger2016pixelwise, kuhn2019plane, xu2019multi, xu2020planar, xu2022multi} and learning-based~\cite{wang2021patchmatchnet, wang2022itermvs, wang2022mvster} methods are considered for comparison. Table~\ref{tab:ETH3D_table} reports the accuracy, completeness and ${F_1}$ score of the point clouds. 

As shown in the table, HPM-MVS can achieve outstanding performance in terms of completeness and ${F_1}$ score among all the methods, and HPM-MVS$_{{\rm{fast}}}$ can outperform in accuracy while guaranteeing completeness and ${F_1}$ score. Notably, HPM-MVS obtains the highest completeness than other methods on both training and test datasets which contain large low-textured areas, because HPM-MVS perceives non-local structured information for recovering geometry.
The visual comparisons are shown in Fig.~\ref{visualize}. It can be clearly observed that our methods produce higher quality point clouds than other competitors, especially in challenging low-textured areas, \eg, red boxes in Fig.~\ref{visualize}.
\begin{table}[]
\caption{$F_1$ score (in $\%$) comparisons of different methods on Tanks \& Temples datasets~\cite{knapitsch2017tanks} at given evaluation threshold.}
    \centering
    \resizebox{0.95\linewidth}{!}{
    \begin{tabular}{c|ccc}
\hline
                                                                        & Method        & Intermediate & Advanced \\ \hline
\multirow{5}{*}{Traditional}                                            & COLMAP~\cite{schonberger2016pixelwise}        & 42.14        & 27.24    \\
                                                                        & PCF-MVS~\cite{kuhn2019plane}       & 55.88        & 34.59    \\
                                                                        & ACMM~\cite{xu2019multi}          & 57.27        & 34.02    \\
                                                                        & ACMP~\cite{xu2020planar}          & 58.41        & 37.44    \\
                                                                        & ACMMP~\cite{xu2022multi}         & 59.38        & 37.84    \\ \hline
\multirow{7}{*}{\begin{tabular}[c]{@{}c@{}}Learning\end{tabular}} & R-MVSNet~\cite{yao2019recurrent}      & 48.40        & 24.91    \\
                                                                        & PatchMatch-RL~\cite{lee2021patchmatch} & 51.81        & 31.78    \\
                                                                        & PatchMatchNet~\cite{wang2021patchmatchnet} & 53.15        & 32.31    \\
                                                                        & IterMVS~\cite{wang2022itermvs}       & 56.22        & 33.24    \\
                                                                        & PVSNet~\cite{xu2022learning}        & 56.88        & 33.46    \\
                                                                        & Effi-MVS~\cite{wang2022efficient}      & 56.88        & 34.39    \\
                                                                        & MVSTER~\cite{wang2022mvster}        & 60.92        & 37.53    \\ \hline
\multirow{2}{*}{Ours}                                                   & HPM-MVS$_{{\rm{fast}}}$ & \textbf{61.59}        & \underline{39.65}    \\
                                                                        & HPM-MVS       & \underline{61.39}        & \textbf{40.80}    \\ \hline
\end{tabular}
    }
\label{tab:TanksAndTemples}
\end{table}

{\bf \noindent{Evaluation on Tanks \& Temples Datasets.}} To further illustrate the robustness of our methods, we test our methods on Tanks \& Temples datasets \textit{without any fine-tuning}. The quantitative results of state-of-the-art methods on both Intermediate and Advanced sets are reported in Table~\ref{tab:TanksAndTemples}. 

\begin{figure*}[h]
        \begin{subfigure}{0.158\linewidth}
		\centering
		\includegraphics[width=1\linewidth]{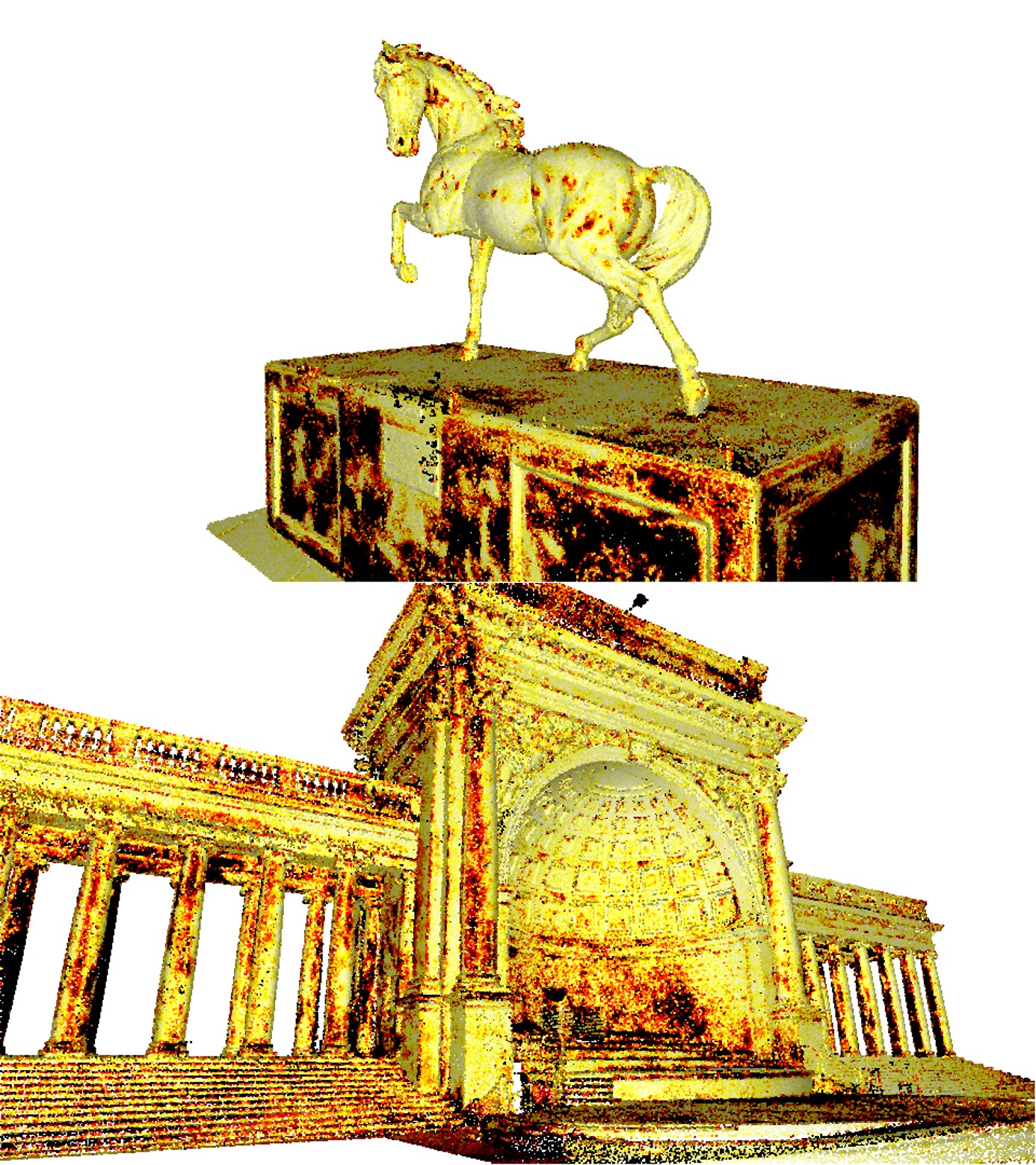}
        \caption{ACMM~\cite{xu2019multi}}
	\end{subfigure}
	\begin{subfigure}{0.158\linewidth}
		\centering
		\includegraphics[width=1\linewidth]{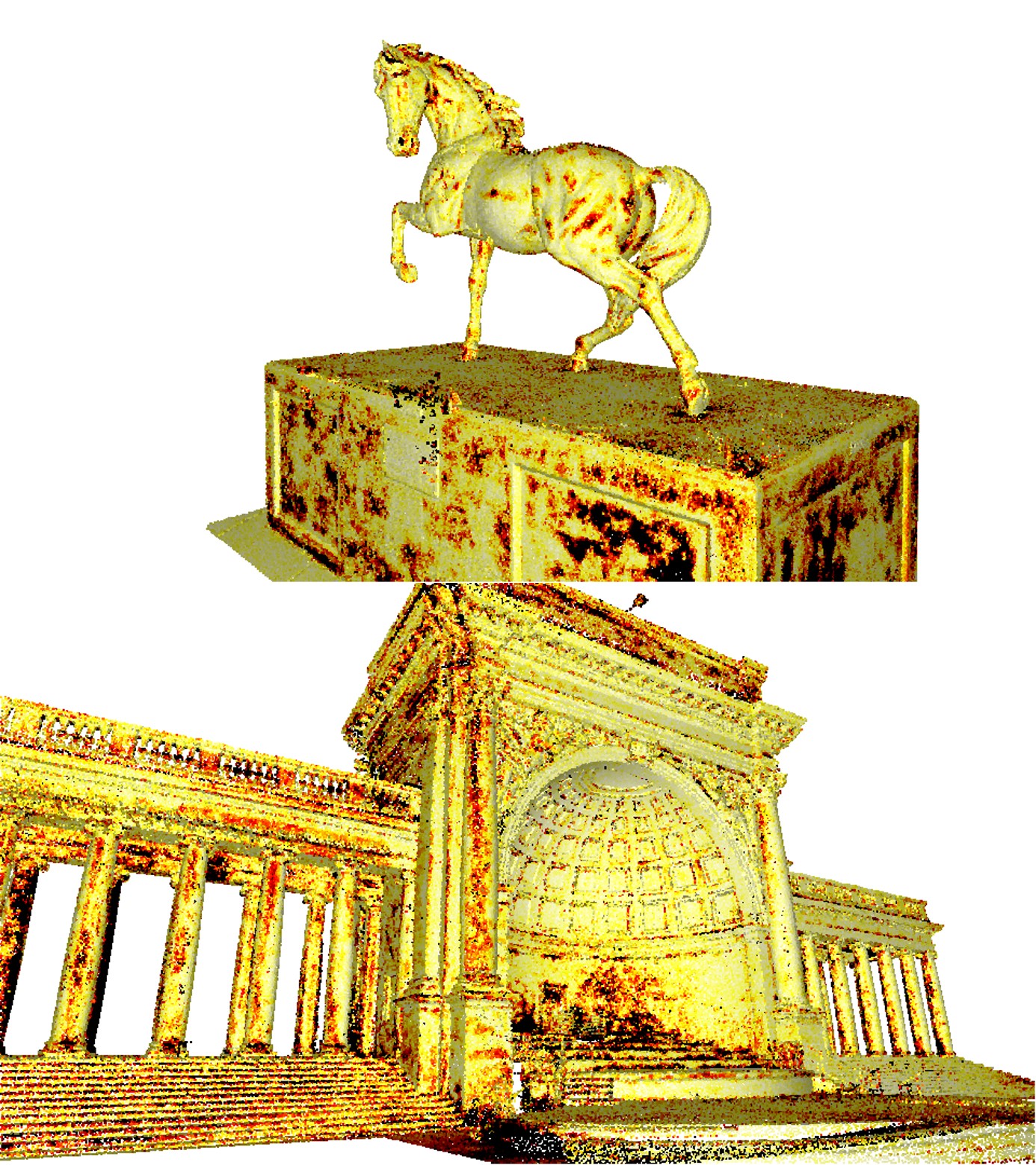}
          \caption{ACMMP~\cite{xu2022multi}}
	\end{subfigure}
        \begin{subfigure}{0.158\linewidth}
		\centering
		\includegraphics[width=1\linewidth]{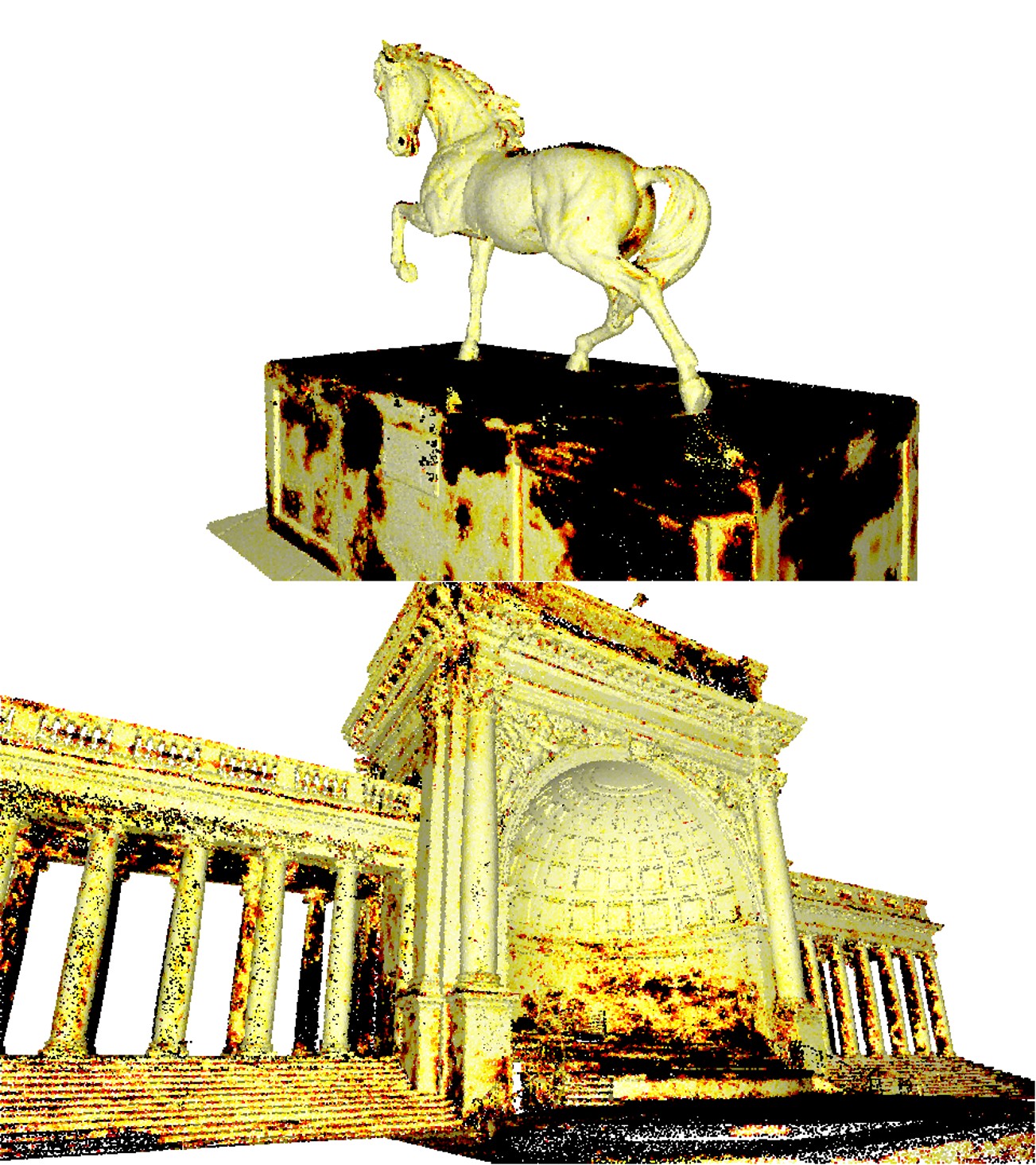}
          \caption{IterMVS~\cite{wang2022itermvs}}
	\end{subfigure}
        \begin{subfigure}{0.158\linewidth}
		\includegraphics[width=1\linewidth]{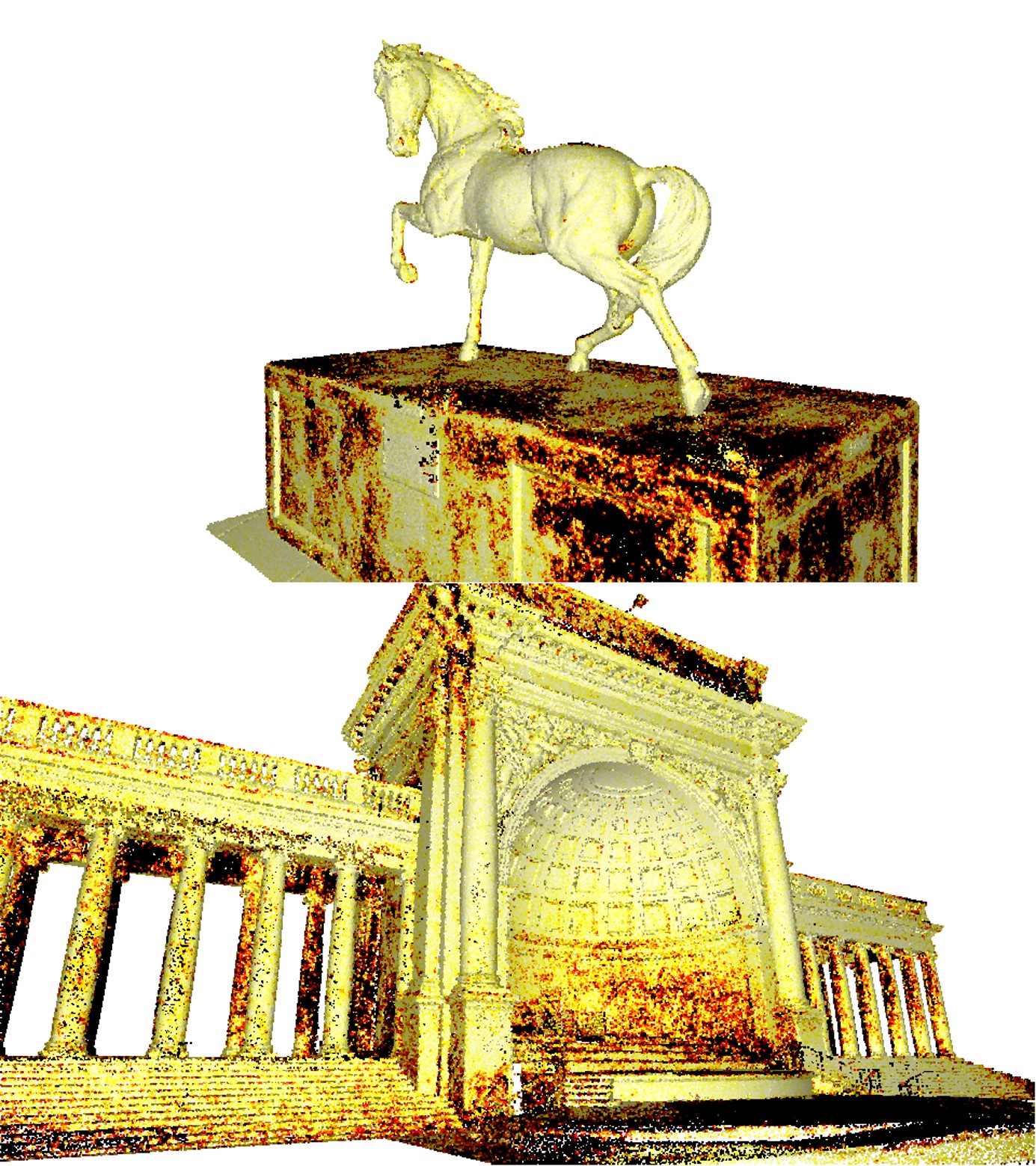}
          \caption{MVSTER~\cite{wang2022mvster}}
	\end{subfigure}
        \begin{subfigure}{0.158\linewidth}
		\includegraphics[width=1\linewidth]{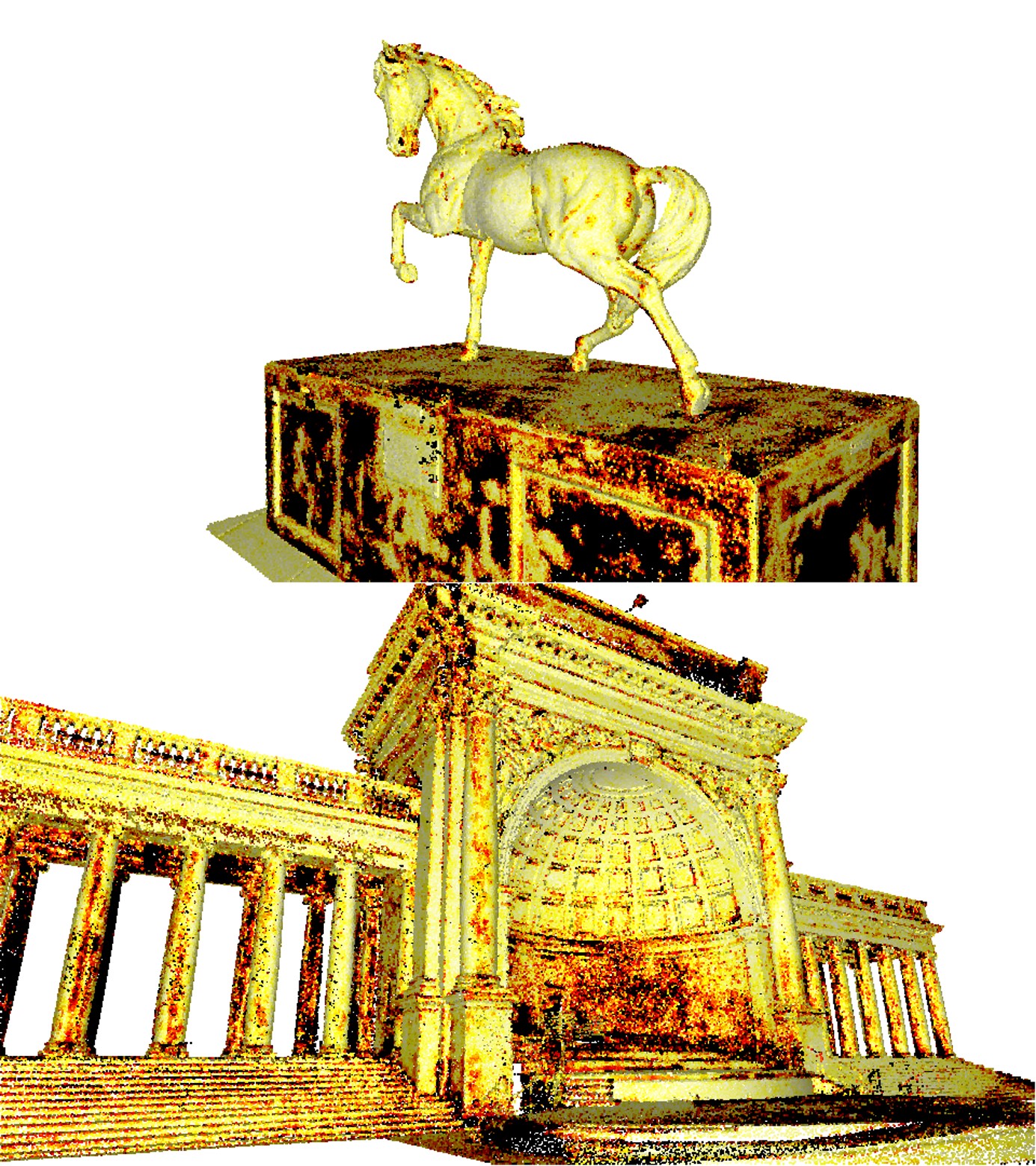}
          \caption{HPM-MVS$_{{\rm{fast}}}$}
	\end{subfigure}
        \begin{subfigure}{0.158\linewidth}
		\includegraphics[width=1\linewidth]{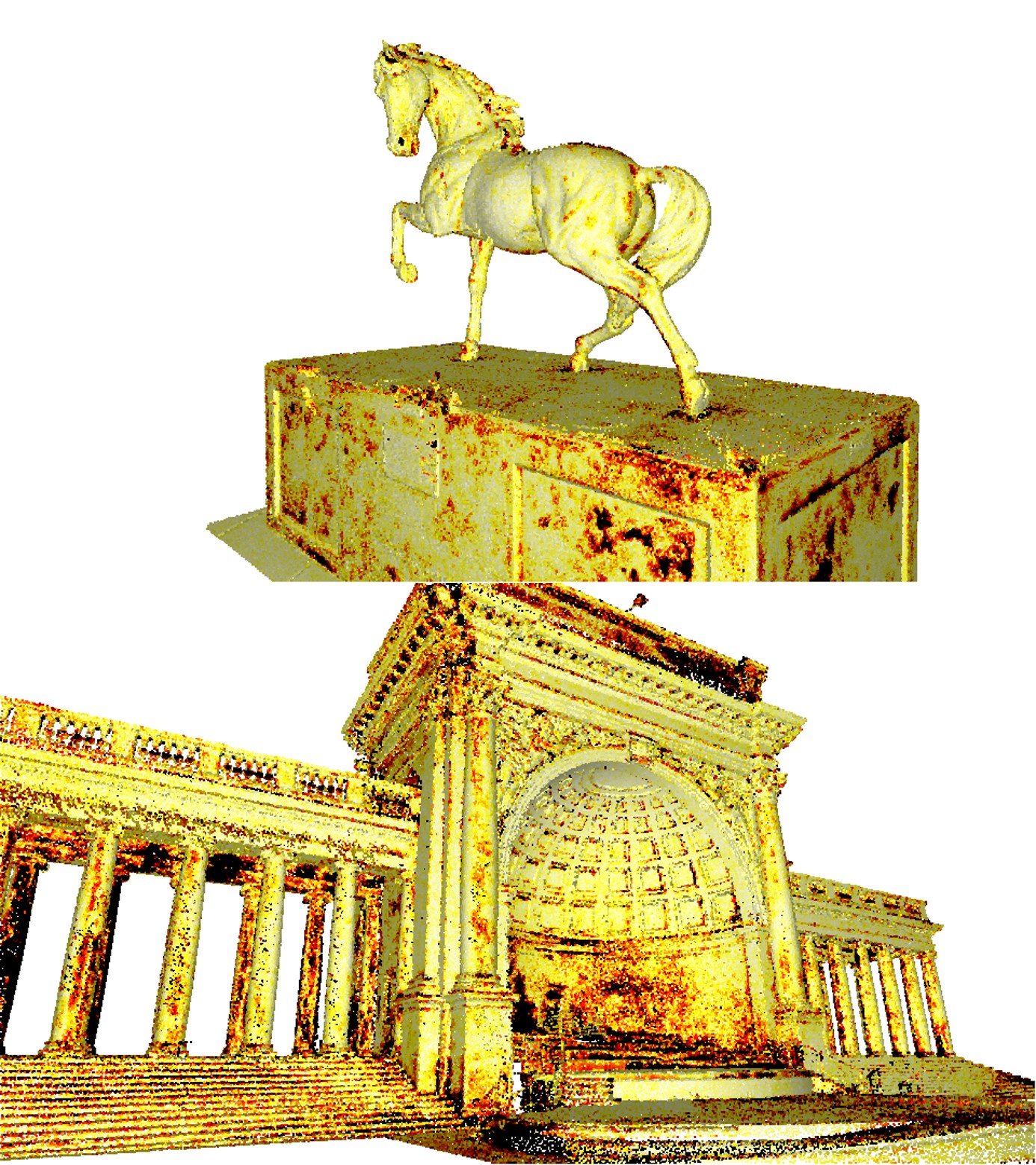}
          \caption{HPM-MVS}
	\end{subfigure}
        \begin{subfigure}{0.02\linewidth}
		\includegraphics[width=1\linewidth]{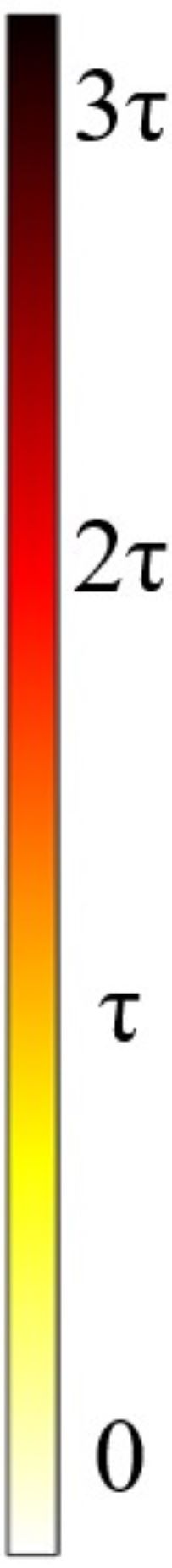}
            \subcaption*{\quad}
	\end{subfigure}
 \centering
	\caption{Qualitative point cloud recall comparisons between MVS methods on Tanks \& Temples datasets~\cite{knapitsch2017tanks} (Horse and Temple). The darker the pixel color is, the greater the error will be. $\tau$ is the recommended threshold for each dataset.}
	\label{Tanks_visualize}
\centering
\end{figure*}

Our methods achieve outstanding performance among all the methods. \textit{Even compared with learning-based methods, our methods still achieve better performance without any data training.} Remarkably, HPM-MVS can greatly improve the performance of Advanced datasets which contain complex details and more low-textured areas difficult to recover. Our method ranks $\bm{1^{st}}$ in Advanced datasets over all the existing traditional MVS methods, and it outperforms the previous best record by \textbf{2.96\%}. This result well confirms that our method can achieve a good balance for MVS reconstruction in details and low-textured areas. Fig.~\ref{Tanks_visualize} presents some visualization examples of point cloud recall comparisons between different methods, which verifies our proposed methods can produce more complete 3D geometries than the others. These obvious advantages suggest that HPM-MVS and HPM-MVS$_{{\rm{fast}}}$ hold powerful generalization ability and can adapt to different application scenarios without any data training.

\subsection{Analysis Experiments}

\begin{table*}[h]
    \caption{Comparative results of our methods under different settings on the training datasets of ETH3D benchmark~\cite{schops2017multi}. N, E, PA, HPM$_{{\rm{fast}}}$ and HPM mean Non-local Sampling Pattern, Extensible Sampling Pattern, planar prior model assistance, Hierarchical Prior Mining framework fast version and Hierarchical Prior Mining framework, respectively.}
    \centering
\resizebox{0.85\linewidth}{!}{
\begin{tabular}{c|ccccc|ccc}
\hline
\multirow{2}{*}{Method} & \multicolumn{5}{c|}{Settings} & \multicolumn{3}{c}{ETH3D train.}                                                                     \\ \cline{2-9} 
                        & N   & E   & PA   & HPM$_{{\rm{fast}}}$  & HPM  & \multicolumn{1}{c|}{$1cm$}               & \multicolumn{1}{c|}{$5cm$}               & $10cm$              \\ \hline
ACMH~\cite{xu2019multi}                    &    &     &     &      &      & \multicolumn{1}{c|}{82.41 / 50.63 / 61.71} & \multicolumn{1}{c|}{96.61 / 74.73 / 83.50} & 98.25 / 82.20 / 88.57 \\
NSP                     & $\checkmark$   &     &     &      &      & \multicolumn{1}{c|}{82.52 / 53.46 / 63.84} & \multicolumn{1}{c|}{\underline{96.63} / 76.15 / 84.52} & \underline{98.46} / 82.40 / 89.19 \\
ESP                     &     & $\checkmark$   &     &      &      & \multicolumn{1}{c|}{82.66 / 51.40 / 62.22} & \multicolumn{1}{c|}{\textbf{96.64} / 75.37 / 84.04} & \textbf{98.47} / 82.09 / 89.03 \\
NESP                    & $\checkmark$   & $\checkmark$   &     &      &      & \multicolumn{1}{c|}{82.47 / 53.98 / 64.22} & \multicolumn{1}{c|}{96.60 / 76.63 / 84.87} & \textbf{98.47} / 82.73 / 89.44 \\
NESP+PA                  & $\checkmark$   & $\checkmark$   & $\checkmark$   &      &      & \multicolumn{1}{c|}{82.88 / \underline{62.37} / \underline{71.32}} & \multicolumn{1}{c|}{96.48 / 84.29 / 88.90} & 97.89 / 88.57 / 92.73 \\
HPM$_{{\rm{fast}}}$                  &     &     & $\checkmark$   & $\checkmark$    &      & \multicolumn{1}{c|}{\textbf{83.32} / 53.53 / 64.66} & \multicolumn{1}{c|}{96.39 / 84.67 / 89.84} & 98.11 / 91.07 / 93.99 \\
HPM                  &     &     & $\checkmark$   &      & $\checkmark$    & \multicolumn{1}{c|}{\underline{83.01} / 62.07 / 70.67} & \multicolumn{1}{c|}{96.06 / \underline{88.42} / \underline{91.97}} & 97.91 / \underline{93.62} / \underline{95.66} \\ \hline
HPM-MVS$_{{\rm{fast}}}$                 & $\checkmark$   & $\checkmark$   & $\checkmark$   & $\checkmark$    &      & \multicolumn{1}{c|}{82.95 / 57.36 / 67.39} & \multicolumn{1}{c|}{96.51 / 86.22 / 90.89} & 98.23 / 92.14 / 94.97 \\
HPM-MVS                 & $\checkmark$   & $\checkmark$   & $\checkmark$   &      & $\checkmark$    & \multicolumn{1}{c|}{82.93 / \textbf{65.16} / \textbf{72.73}} & \multicolumn{1}{c|}{96.13 / \textbf{89.98} / \textbf{92.88}} & 97.97 / \textbf{94.59} / \textbf{96.22} \\ \hline
\end{tabular}
}

    \label{tab:abaltion study}
\end{table*}
{\bf \noindent{Ablation Study.}} In this section, we conduct an ablation study on the multi-view high-resolution training datasets of ETH3D benchmark to validate the performance of different parts in our proposed methods. We divide them into five sections, which are Non-local Sampling Pattern, Extensible Sampling Pattern, planar prior model assistance, Hierarchical Prior Mining framework fast version and Hierarchical Prior Mining framework. The baseline method is ACMH\footnote{Note that, we combine ACMH with geometric consistency here.}~\cite{xu2019multi}. Table~\ref{tab:abaltion study} summarizes the effectiveness of each individual part in our proposed methods.

For the basic MVS method with NESP, we remove the no-local part and the extensible part, respectively. The results show that these two proposals can both enhance the completeness and ${F_1}$ score. In particular, the improvement of the non-local strategy is more obvious than the extensible strategy. In some circumstances, especially in low-textured areas, several severe errors always cause the hypothesis propagation into local optimal solutions. Hence, the non-local operation can effectively prevent this situation from getting worse.

Based on NESP, the planar prior model constructed at the current scale generates a potential hypothesis for each pixel in the image. As for two versions of the HPM framework, one can make several observations from the results: 1) HPM and HPM$_{{\rm{fast}}}$ achieve outstanding performance at relatively high thresholds (\eg, $5cm$ and $10cm$), because they have a strong focus on sizable low-textured areas. 2) We notice that HPM$_{{\rm{fast}}}$ will suffer some loss of completeness at low thresholds (\eg, $1cm$), because it obtains the prior model at the lowest resolution scale which may lead to ambiguities in details. 3) The method with NESP as the basis for subsequent processes can provide a set of reliable points for planar prior model construction.

\begin{table}[]
    \centering
    \caption{Generalization performance of NESP on ETH3D high-resolution training datasets~\cite{schops2017multi}.}
    \label{tab:generalization}
    \resizebox{0.9\linewidth}{!}{
    \begin{tabular}{c|ccccc}
\hline
\multirow{2}{*}{Method}     & \multicolumn{5}{c}{ETH3D train.}      \\
                            & $1cm$   & $2cm$   & $5cm$   & $10cm$  & $20cm$  \\ \hline
ACMM~\cite{xu2019multi}                        & 67.58 & 78.86 & 87.68 & 91.70 & 94.41 \\
ACMP~\cite{xu2020planar}                        & 68.72 & 79.79 & 88.32 & 92.03 & 94.43 \\
ACMMP~\cite{xu2022multi}                       & 71.57 & 83.42 & 92.03 & 95.54 & 97.37 \\ \hline
\multirow{2}{*}{ACMM+NESP}  & 70.70 & 81.01 & 88.80 & 92.26 & 94.55 \\
                            & \textcolor[rgb]{0,0.69,0.314}{3.12$\uparrow$}  & \textcolor[rgb]{0,0.69,0.314}{2.15$\uparrow$}  & \textcolor[rgb]{0,0.69,0.314}{1.12$\uparrow$}  & \textcolor[rgb]{0,0.69,0.314}{0.56$\uparrow$}  & \textcolor[rgb]{0,0.69,0.314}{0.14$\uparrow$}  \\ \hdashline[2.5pt/5pt]
\multirow{2}{*}{ACMP+NESP}  & 70.87 & 81.45 & 89.43 & 92.72 & 94.78 \\
                            & \textcolor[rgb]{0,0.69,0.314}{2.15$\uparrow$}  & \textcolor[rgb]{0,0.69,0.314}{1.66$\uparrow$}  & \textcolor[rgb]{0,0.69,0.314}{1.11$\uparrow$}  & \textcolor[rgb]{0,0.69,0.314}{0.69$\uparrow$}  & \textcolor[rgb]{0,0.69,0.314}{0.35$\uparrow$}  \\  \hdashline[2.5pt/5pt]
\multirow{2}{*}{ACMMP+NESP} & 74.54 & 85.33 & 93.25 & 96.45 & 97.99 \\
                            & \textcolor[rgb]{0,0.69,0.314}{2.97$\uparrow$}  & \textcolor[rgb]{0,0.69,0.314}{1.91$\uparrow$}  & \textcolor[rgb]{0,0.69,0.314}{1.22$\uparrow$}  & \textcolor[rgb]{0,0.69,0.314}{0.91$\uparrow$}  & \textcolor[rgb]{0,0.69,0.314}{0.62$\uparrow$}  \\ \hline
\end{tabular}
    }
    
\end{table}

{\bf \noindent{Generalization Performance of NESP.}}
In addition, we perform a more extensive comparison on the multi-view high-resolution training datasets of ETH3D benchmark to show the generalization ability of NESP. Several kinds of state-of-the-art diffusion-like propagation methods are integrated with NESP for evaluation. The considered methods are ACMM~\cite{xu2019multi}, ACMP~\cite{xu2020planar} and ACMMP~\cite{xu2022multi}. Note that ACMMP is a very recent state-of-the-art geometric-only method.  Each method is tested under different thresholds, and results are shown in Table~\ref{tab:generalization}.

Impressively, NESP dramatically improves the ${F_1}$ score for all tested methods at relatively low thresholds, \eg, $1cm$ and $2cm$. The improvement of NESP working with multi-scale strategy is more obvious, because the non-local and extensible strategy works even better at low-resolution scale. Notably, the combination of NESP and ACMMP outperforms all the competitors and achieves state-of-the-art ${F_1}$ score of \textbf{85.33\%} / \textbf{96.45\%} at $2cm$ / $10cm$ thresholds. Based on the results, the following conclusions can be drawn: 1) NESP is a generalized module that can be adapted to other MVS methods and boost their performance. 2) Combined with the multi-scale strategy, NESP can achieve better performance. 

More analysis experiments and point cloud visualizations can be found in the supplementary.

\section{Conclusion}
In this paper, we presented an HPM-MVS method which can efficiently perceive non-local structured information to achieve high-quality reconstruction. Based on the non-local and extensible strategy, we first propose our NESP module. Focusing on the planar prior construction in marginal regions, we employ KNN to search non-local credible points and obtain potential hypotheses. Further, we design an HPM framework which explores planar prior at different scales to assist MVS. Through extensive experiments, we have demonstrated the rationality of each key component of our method and its superiority against existing methods.
In the future, we plan to combine our method with a probabilistic graphical model to further improve its ability to handle the reconstruction of details and low-textured regions.

{\small
\bibliographystyle{ieee_fullname}
\bibliography{egbib}
}

\end{document}